\documentclass[11pt,oneside,letter]{article}
\usepackage[top=1 in, bottom=1 in, left=0.85 in, right=0.85 in]{geometry}

\usepackage{amsmath,amsfonts,bm}

\def\eqref#1{equation~\ref{#1}}

\def\1{\bm{1}}

\def\eps{{\epsilon}}

\def\vmu{{\bm{\mu}}}
\def\vtheta{{\bm{\theta}}}

\def\ve{{\bm{e}}}

\def\vg{{\bm{g}}}

\def\vi{{\bm{i}}}

\def\vv{{\bm{v}}}

\def\mP{{\bm{P}}}

\DeclareMathAlphabet{\mathsfit}{\encodingdefault}{\sfdefault}{m}{sl}
\SetMathAlphabet{\mathsfit}{bold}{\encodingdefault}{\sfdefault}{bx}{n}

\def\gE{{\mathcal{E}}}
\def\gF{{\mathcal{F}}}

\def\gO{{\mathcal{O}}}

\def\sA{{\mathbb{A}}}

\def\sP{{\mathbb{P}}}

\def\sR{{\mathbb{R}}}
\def\sS{{\mathbb{S}}}

\newcommand{\E}{\mathbb{E}}

\newcommand{\lr}{\alpha}

\usepackage{float}
\usepackage{mathtools}

\usepackage[round]{natbib}
\usepackage{makecell}
\usepackage{amsmath,amssymb,graphicx,url}
\usepackage{thmtools,thm-restate,wrapfig,enumitem,mathabx}
\usepackage{booktabs}
\newtheorem{assumption}{Assumption}
\newtheorem{theorem}{Theorem}

\newtheorem{lemma}{Lemma}
\newtheorem{corollary}{Corollary}

\usepackage{color}
\usepackage{subfigure}
\usepackage{colortbl}
\usepackage{multirow}

\usepackage{bbm}

\usepackage[utf8]{inputenc} 
\usepackage[T1]{fontenc}    
\usepackage{hyperref}       
\usepackage{url}            
\usepackage{booktabs}       
\usepackage{amsfonts}       
\usepackage{nicefrac}       
\usepackage{microtype}      
\usepackage{xcolor}         

\title{Zeroth-Order Policy Gradient for Reinforcement Learning from Human Feedback without Reward Inference}
\date{}
\author{
	Qining Zhang \\
	University of Michigan, Ann Arbor \\  \texttt{qiningz@umich.edu} \\ 
	\and
	Lei Ying \\
	University of Michigan, Ann Arbor \\  \texttt{leiying@umich.edu}\\
	\\
}

\usepackage{times}
\usepackage{hyperref}
\usepackage{url}
\usepackage{bbm}
\usepackage[ruled,vlined,linesnumbered]{algorithm2e}
\usepackage{graphicx}
\usepackage[round]{natbib}

\SetKwInOut{Parameters}{Parameters}
\newcommand{\trim}{\mathrm{trim}}

\date{}

\begin{document}

\maketitle

\begin{abstract}
Reward inference (learning a reward model from human preferences) is a critical intermediate step in the \emph{Reinforcement Learning from Human Feedback} (RLHF) pipeline for fine-tuning \emph{Large Language Models} (LLMs). In practice, RLHF faces fundamental challenges such as distribution shift, reward model overfitting, and problem misspecification. An alternative approach is direct policy optimization without reward inference, such as \emph{Direct Preference Optimization} (DPO), which provides a much simpler pipeline and has shown empirical success in LLM applications. However, DPO utilizes the closed-form expression between the optimal policy and the reward function, which is only suitable under the bandit setting or deterministic MDPs. This paper develops two RLHF algorithms without reward inference for {\em general} RL problems beyond bandits and deterministic MDPs, and {\em general} preference models beyond the Bradley-Terry model. The key idea is to estimate the local value function difference from human preferences and then approximate the policy gradient with a zeroth-order gradient approximator. For both algorithms, we establish polynomial convergence rates in terms of the number of policy gradient iterations, the number of trajectory samples, and human preference queries per iteration. Numerical experiments in stochastic environments validate the performance of our proposed algorithms, outperforming popular RLHF baselines such as DPO and PPO. Our paper shows there exist provably efficient methods to solve general RLHF problems without reward inference. 
\end{abstract}

\section{Introduction}
\begin{figure}
    \centering
    \includegraphics[width=0.85\linewidth]{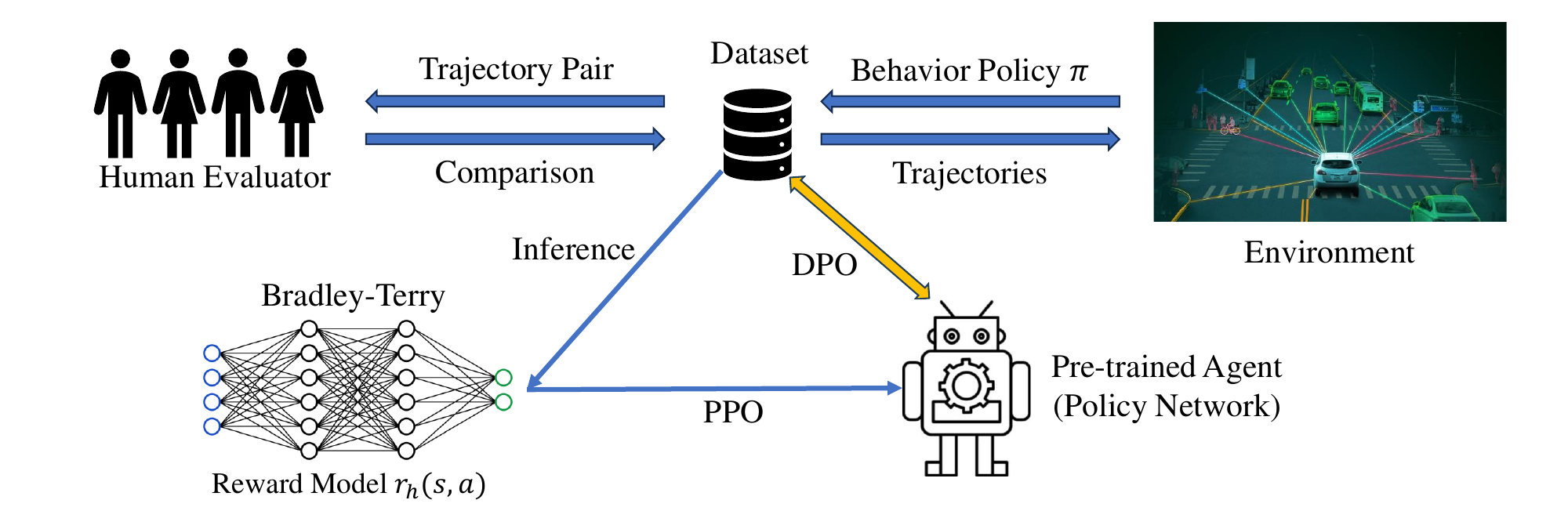}
    \caption{A diagram illustrating classic policy-based RLHF and DPO: classic RLHF involves three steps: (i) policy pre-training: pre-train a policy network (agent), (ii) reward inference: collect trajectories from the environment using a behavior policy, query the human comparison for each trajectory pair and train a reward neural network through maximizing the likelihood under the Bradley-Terry model, and (iii) policy training with reward model: train the policy network with reward signals sampled from the reward network. DPO does not train a reward network but directly optimizes the policy network from human preferences.}
    \label{fig:PPO}
\end{figure}

In the past decade, we have witnessed unprecedented success in applying \emph{Reinforcement Learning} (RL) to many applications, such as video games~\citep{knox08gameAI,warnell18gameAI}, recommendation and search~\citep{zeng16recommendation,kohli13recommendation}, autonomous driving~\citep{kiran21driving}. RL studies the interaction between decision-making agents and an evolving dynamic environment. At each time step, the agent takes a certain decision (action) given the current state, a reward signal to measure the quality of that decision is provided by the environment. The agent's goal is to learn a policy to maximize the cumulative reward, and the quality of the learned policy will depend on the per-step reward function. In classic RL, this reward function is usually handcrafted by domain experts to ensure it aligns with human interests. However, the problem of identifying a ``good'' reward function, also referred to as \emph{Inverse Reinforcement Learning} (IRL), is non-trivial and one of the most fundamental problems in the history of RL~\citep{Ng00IRL}. In recent years, \emph{Reinforcement Learning from Human Feedback} (RLHF) that uses human preference feedback as a signal to recover a reward function has emerged to fine-tune \emph{Large Language Models} (LLMs), which has delivered significant success~\citep{Christiano17RLHFTRPO,wu21llm,nakano21llm,ziegler19llm,stiennon20llm, Ouyang22InstructGPT}. RLHF follows the diagram shown in Fig.~\ref{fig:PPO}, which includes three major steps~\citep{Ouyang22InstructGPT}: (i) pre-train a policy neural network, (ii) collect sets of trajectory pairs and query human evaluators for preferences over trajectory pairs to train a reward model using maximum likelihood to align with human feedback, and (iii) use policy-optimization-based RL algorithms such as PPO~\citep{schulman2017ppo} to fine-tune the policy network with the reward signals generated by the reward model. The reward inference intermediate step, which trains the reward network, is crucial to obtaining a high-quality policy through RL in the final step.

\textbf{Drawbacks of Reward Inference.} 
To train a good reward model, i.e., to infer the underlying per-step reward function from human feedback~\citep{Christiano17RLHFTRPO,Wang23RLHFP2R}, the most common approach is to assume the feedback is generated based on a preference model such as the \emph{Bradley-Terry} model~\citep{Bradley52Preference}, and then maximize the log-likelihood of the collected trajectory comparison dataset accordingly over all possible (parameterized) reward functions. This procedure is indeed analyzed in most theoretical RLHF papers for both offline~\citep{zhu23RLHF,Zhan24RLHFOffline} and online settings~\citep{saha23duelingrl,Zhan24RLHF,wu24PbRL,Wang23RLHFP2R,Du24RLHFPG}. However, several challenges occur in practice for reward model training such as double problem misspecification, reward model evaluation without ground truth, distribution shift, and overfitting in joint reward model and policy training \citep{CasDavShi_23}. These drawbacks are also reflected in the theoretical results, e.g., overfitting of the maximum likelihood estimator (MLE) in \citet{zhu24RLHFoverfit}. Moreover, similar to the dilemma in IRL, the reward function that could explain human feedback is often not unique, especially when given limited training trajectories~\citep{Arora21IRL,Ng00IRL}. Some reward models may make it difficult for agents to learn a good RL policy.

\textbf{DPO.} To avoid the drawbacks of the reward inference in RLHF, \citet{Rafailov23DPO} proposed an algorithm called \emph{Direct Preference Optimization} (DPO) which fine-tunes the LLM \emph{directly} from human preferences. Based on the Bradley-Terry preference model and a closed-form expression of the optimal policy given a reference policy and the reward function, DPO constructs a loss function directly from human feedback for learning the optimal policy to avoid reward inference.
This provides a much simpler pipeline and has great empirical performance~\citep{Rafailov23DPO,rafailov24dpo,rafailov24dpoQ}. However, 
the closed-form expression of the optimal policy that DPO builds on is only for non-parametric policies and its theoretical justification only works for the bandit setting \citep{Rafailov23DPO} or RL problems with deterministic transitions \citep{rafailov24dpoQ}. It remains an open question how to solve general RLHF problems without reward inference.

\textbf{RLHF without Reward Inference.} 
Recently, value-based RLHF algorithms without global reward inference have been theoretically developed and analyzed~\citep{xu20RLHFQ,zhang24RLHFQ} based on a dueling bandit approach~\citep{Bengs21DuelingBanditSurvey}. The results, however, only hold for MDPs in tabular settings with finite state and action spaces. \citet{chen22RLHF} studied the function approximation regime, 
but their algorithm requires both the true preference model and the transition kernel to belong to a known function class, which is also far from practice. The result also depends on the function class complexity, which is usually large for most function approximators in practice. So far, no provable policy-based algorithm in this category has been developed.

This paper addresses the following important question:

\begin{quote}
    \em Does there exist a provably efficient RLHF approach that does not require a reward model and works for general RL problems such as stochastic MDPs or infinite state and action spaces? 
\end{quote}

\subsection{Main Contributions}

DPO~\citep{Rafailov23DPO} establishes a direct connection between human preferences and RL based on the Bradley-Terry model and the 
optimal policy in closed form:
\begin{equation}
    \pi^*(a|x) \propto \pi_{\hbox{\scriptsize ref}}(a|x)\exp\left(\frac{1}{\beta}r(x,a)\right),\label{eq:opt}
\end{equation} where $r(x,a)$ is the reward in state $x$ with action $a,$ $\pi_{\hbox{\scriptsize ref}}$ is a reference policy and $\pi^*$ is the optimal policy. Based on the direction connection, the policy optimization can be formulated as a direct matching between human preference and the optimal policy with a log-likelihood loss function. 
In a recent paper \citep{rafailov24dpo}, it has been further shown that DPO solves a KL-divergence-constrained policy optimization problem for the {\em deterministic} token-level MDP for LLMs, where the next state is deterministic given the current state and action. 
For general RL problems with parameterized policies, \eqref{eq:opt} does not hold, and it is often hard if not impossible to obtain a ``global'' function like it that connects the optimal policy and the reward (hence human feedback). 

This paper exploits the ``local'' relation between human feedback and policy optimization. In particular, given a policy $\pi_\vtheta$ and a perturbed version of the policy $\pi_{\vtheta+\vv}$ where $\vv$ is a small perturbation vector, we use human feedback over the trajectories generated from both policies to inform the direction of a more preferred policy. Intuitively, if one trajectory is preferred over the other, the policy that generates this trajectory is likely to have a higher value. Then given a preference model such as the Bradley-Terry model, we can further estimate the value function differences of the two policies, $V(\pi_{\vtheta+\vv})-V(\pi_\vtheta)$, where $V(\pi)$ is the value function associated with policy $\pi.$
Finally, the value difference can be used as an estimator of policy gradient, $ \nabla_{\vtheta} V(\pi_{\vtheta})$, following the zeroth-order optimization approach~\citep{nesterov17zosgd,Ghadimi13ZOSGD} to improve the policy. 

Based on this idea, this paper proposes two RLHF algorithms without reward inference: \emph{\textbf{Z}eroth-Order \textbf{P}olicy \textbf{G}radient} (ZPG) and \emph{\textbf{Z}eroth-Order \textbf{B}lock-\textbf{C}oordinate \textbf{P}olicy \textbf{G}radient} (ZBCPG), both from Human Feedback. 
ZBCPG differs from ZPG in its policy perturbation rule, which has lower computational complexity and allows parallel optimization since one can sample multiple perturbed policies to perform policy gradient and aggregate the estimated gradient. Under mild assumptions, both algorithms have the following rate of convergence to a stationary policy:  
$$\gO\left(\frac{Hd}{T} + \frac{d^2 \sqrt{\log M}}{\sqrt{M}} + \frac{Hd\sqrt{d}}{\sqrt{N}}\right),$$ 
where $d$ is the dimension of policy network parameter $\vtheta,$ $H$ is the planning horizon, $T$ is the number of policy gradient steps, $N$ is the number of policy perturbations each step, and $M$ is the number of human queries for each pair of trajectories. 

We remark that~\citet{tang24zohf} proposes a similar approach towards utilizing human feedback and a zeroth-order gradient descent algorithm from ranking data. However, they assume an error-free ranking oracle over policies based on their value functions, which makes their problem a deterministic optimization problem and does not apply to trajectory preference data like in RLHF and DPO.  
This paper studies RLHF with trajectory preferences and quantifies the impacts of stochastic trajectories and human preferences on the rate of convergence of RLHF without reward inference.

\subsection{Related Work}\label{sec:relatedwork}
We review recent developments in RLHF and zeroth-order optimization for both empirical and theoretical results. A thorough survey on RLHF can be found in \citet{kaufmann23RLHFsurvey} and \citet{CasDavShi_23}

\textbf{Empirical Studies of Direct RLHF in LLMs.}
DPO~\citep{rafailov24dpo} and SLiC-HF~\citep{zhao23RLDHF} have empirically shown it is possible to directly learn an RL policy from human preference. In DPO, the authors solve a KL-divergence-constrained reward maximization problem for a prompt response generation problem similar to the contextual bandit. The optimal policy for this problem has a closed-form expression, and the reward of each prompt-response pair can be computed knowing the optimal policy and the reference policy. The authors then minimize a log-likelihood loss by plugging the reward expression into the Bradley-Terry model to measure the alignment between the policy and the human preference without training a reward network.~\citet{rafailov24dpoQ} extends this loss to token-level MDP on the condition that the MDP transition is deterministic, i.e., the next state is a concatenation between the current state and action. This limits the DPO loss minimization approach to LLM problems. In general MDPs with stochastic transition, the DPO loss cannot be computed following the derivation in~\citep{rafailov24dpoQ}.~\citet{azar24phiPO} extends DPO to a wider class of RL problems, even avoiding the notion of an underlying reward function. Instead of maximizing the reward in a KL-constrained problem like DPO, the authors proposed to optimize a general non-decreasing function of the ground-truth population-level preference probability. Other variants are also studied.~\citet{ethayarajh24kto}
considers aligning policy with humans and designs loss from a prospect theory perspective, and~\citet{tang24GPO} considers optimizing a general loss of the preference loss instead of the log-likelihood.~\citet{dong24onlineDPO} and~\citet{xiong24iterativeDPO} proposed to obtain human feedback in an online fashion to mitigate the distribution-shift and over-parameterization phenomenon. Attempts to understand the theoretical performance of RLDHF algorithms such as DPO are made in~\citep{azar24phiPO}, but the authors only showed the existence of optima of the loss function, without any policy optimality and sample complexity guarantees.

\textbf{RLHF with Provable Sample Complexity.} Two major approaches have been studied to learn the optimal policy from human preference data. The first is similar to the traditional RLHF paradigm, such as PPO used in empirical studies, which infers a reward function, or sometimes a utility function, and then trains an RL agent. This reward inference step is also called surrogate learning in the preference-based RL literature, i.e., see \citep[Sec. 2.4]{wirth17pbrlsurvey}. The second approach directly optimizes the policy from human preferences. Empirical algorithms for both approaches have been developed without theoretical guarantees for a few years. For example, reward function or utility function estimation followed by an RL policy search algorithm has been proposed and studied in~\citep{Akrour14PbRL,Wirth16PbRL,Christiano17RLHFTRPO}, while direct policy learning from humans through trajectory sampling also has been proposed in using a heuristic evolutionary algorithm \citep{busa14PbRLevolution,akrour11PbRLevolution}, or from a Bayesian Markov chain Monte Carlo perspective~\citep{wilson12PbRLMCMC}. However, it was not until recently that algorithms with provable theoretical guarantees were proposed. From the reward inference approach, \citet{novoseller20rlhf} took a Bayesian perspective and maintained a posterior distribution over the reward function and the transition kernel, which is computationally costly.
\citet{Wang23RLHFP2R} assumes the reward function can be linearly parameterized with known feature embedding and proposes a preference-to-reward interface (a reward model) using an online policy update and a baseline trajectory for comparison. The authors theoretically showed that for a general known preference model setting, RLHF is no harder than RL. A similar analysis framework is adopted in contemporary theoretical works, e.g., \citep{saha23duelingrl,Zhan24RLHFOffline, Zhan24RLHF,zhu23RLHF,Kong22RLHF,wu24PbRL}, for both online, offline, and hybrid RL problems, where the algorithms first learn the linear parameter of the reward function and perform value-based RL on the learned reward function. The analysis first characterizes the error of the reward parameter using the concentration of the MLE estimator and then propagates this error to the value-based RL algorithm. Specifically, \citet{saha23duelingrl} considers the tabular RL setting and assumes a known feature embedding for each trajectory with a known transition kernel. Then, the agent is directed by RL to explore the feature direction where the uncertainty in the reward function is large. \citet{wu24PbRL} extends the scenario to linear MDPs with unknown transition and uses least square value iteration to solve the RL problem. \citet{zhu23RLHF} extends the work \citep{saha23duelingrl} to the offline RL scenario where a pessimistic estimator of the reward parameter is used to combat the insufficient data coverage in offline settings. \citet{Zhan24RLHF} replaces the linear reward parameterization with a general function class under the realizability condition. This also enables them to solve the unknown preference model as long as it is in the known function class. \citet{Zhan24RLHF} and \citet{Kong22RLHF} study the hybrid RL problem with human preference, where they first use an exploratory policy from an optimal design perspective to improve the coverage of the offline dataset and to extract more information useful for reward inference. Then, the problem will be solved under the general RL framework. All papers above took a value-based approach in RL and \citet{Du24RLHFPG} analyzed natural policy gradient with reward inference. Direct policy learning from human preference has been less analyzed compared to reward inference approaches. In tabular MDPs, \citet{xu20RLHFQ} and \citet{zhang24RLHFQ} reduce the RL problem to a sequence of dueling bandit problems. However, the approach is only suitable for MDPs with finite state and action spaces. For general MDPs, \citet{chen22RLHF} first uses function approximation for the mapping from trajectory pair to human preference, and learns the RL transition kernel from a least square estimator with this preference approximator. The optimal policy can then be learned using a dueling bandit approach. However, their results assume the true preference model and transition kernel are inside the known function class with small complexity, which is strong in real applications. RLHF has also been studied in other aspects. \citet{li24rlhf} studies RL from the human behavior dataset from a dynamic choice perspective. ~\citet{chakraborty24RLHFBilevel} formulated the reward inference and the policy optimization as a bilevel optimization problem, and~\citet{kausik24RLHFPOMDP} studies RLHF with partially observed rewards and states. \citet{zhu24RLHFoverfit} studies the overfitting issue in reward model training. Recently, \citet{xie2024xpo} combined DPO with optimistic exploration to design XPO in the function approximation regime with provable convergence. However, their algorithm and results are still dependent on the DPO loss, which only works for deterministic MDPs.

\textbf{Zeroth-Order Optimization and Evolutionary Strategies.} The zeroth-order optimization problem has been studied in the convex and non-convex optimization literature for more than a decade~\citep{Ghadimi13ZOSGD,nesterov17zosgd}, where the stochastic gradient descent algorithm with two-point gradient estimator is most widely used. The convergence rate in smooth functions is first studied in~\citep{Ghadimi13ZOSGD} in both convex and non-convex settings. In the convex setting, the algorithm finds the optimal point while in the non-convex setting, the algorithm finds a stationary point. The rate of convergence for non-convex functions is improved in~\citet{nesterov17zosgd}. Variants of stochastic gradient descent such as variance reduction techniques~\citep{liu18ZOSVRG} and ADMM~\citep{liu18ZOADMM,Gao18ZOSGDsmoothing} have also been studied in the zeroth-order literature.~\citep{cai21zobcd} extends the zeroth-order method to blocked coordinate descent for computational efficiency and~\citet{liu18zosignsgd} extends the method to analyze a signed version of SGD which is more memory efficient in federated settings since each element of the gradient takes only one bit. In recent years, the study of evolutionary strategies~\citep{rechenberg1973evolution} brought zeroth-order optimization methods to optimize RL agents, which has achieved empirical success as a scalable alternative to classic RL algorithms such as Q-learning or policy gradient~\citep{salimans2017evolution, conti2018evolution}. This strategy has also been applied in preference-based RL as well~\citep{busa14PbRLevolution,akrour11PbRLevolution}. However, the theoretical guarantees and provable algorithms remain under-explored. The zeroth-order optimization technique has been proposed in optimizing LLMs~\citep{Malladi23ZOLLM,zhang2024ZOLLM}, but they only implement it in the procedure of policy optimization from the reward network to avoid the heavy memory burden in back-propagation. In their studied problem, the loss function can be explicitly estimated or calculated, and thus can be queried to construct the zeroth-order gradient, which is more similar to classic evolutionary strategies. Our work is different in that the loss function, i.e., the value function, cannot be directly queried without reward feedback, so the zeroth-order algorithms cannot be directly used. However, we view human feedback as a natural source of zeroth-order information and apply the method directly from human preference.

\textbf{Preference Models.} The study of modeling the rationale of human decision-making has been pertinent for almost a century. The most popular and widely studied model in the literature is the random utility model (RUM) in social choice theory~\citep{azari2012randomutility}, which was first developed as early as 1920s~\citep{thurstone1927probit}. In RUM, for each decision-maker, each choice is associated with a utility that consists of two portions: a shared public utility same for all people, and a private utility (noise) unique to each person. People are modeled as utility maximizers and therefore will choose the choice that has the largest total utility. The private utility is assumed to follow a certain distribution among the population of decision-makers. It is not hard to see that the public utility will determine which choice is more preferred among the population, and the distribution of the private utility will determine how much it is more preferred, i.e., the population preference probability. Different distribution assumptions give rise to different preference models with different link functions. If the private utility follows a normal distribution, the preference model is called the Probit model where the link function is the cumulative distribution function of a normal distribution. The Probit model is the first proposed and studied preference model in the literature~\citep{thurstone1927probit}. Suppose the private utility follows the Gumbel distribution. We recover the Bradley-Terry model, also known as the logit model, with the logistic link function. Other preference models include the linear model with linear link function, the Weibull model, the Cauchy model, and the complementary log-log model, which have been studied in the literature for specific applications, e.g., see~\citep{train2009choice,greene2000econometric,greene2010model} for a comprehensive review. In general, when the utility is close among choices, the Bradley-Terry model and the Probit model are often the most accurate and easy to implement.

\section{Preliminaries}
In this section, we introduce the notations and preliminaries of our paper. We first introduce the RL problem with general trajectory reward. Then, we describe the policy parameterization, and the human preference feedback mechanism, as well as our optimization problem formulation.

\textbf{Episodic RL:} We consider an episodic RL problem $\mathcal{M} = (\sS, \sA, H, \mP, \vmu_0)$, where $\sS$ is the state space and $\sA$ is the action space (both can be continuous), $H$ is the RL planning horizon, $\mP = \{\mP_h\}_{h=1}^H$ is the set of transition kernels, and $\vmu_0$ is the initial distribution. At the beginning of each episode, the agent will choose a policy $\pi$ represented by $H$ functions $\{\pi_h:\sS \to \mathcal{P}(\sA)\}_{h=1}^H$, where $\mathcal{P}(\sA)$ denotes the set of all probability distributions over the action space. Then, an initial state $s_1$ is sampled from the initial distribution $\vmu_0$. At step $h$, the agent takes an action $a_h = \pi_h(s_h)$ after observing state $s_h$. The environment then moves to the next state $s_{h+1}$ sampled from the distribution $\mP_h(\cdot|s_h,a_h)$ without revealing any reward feedback. We use $\tau$ to denote a trajectory with planing horizon $H$, i.e., $\tau = \{(s_h, a_h)\}_{h=1}^H$.

\textbf{Trajectory Reward:} we assume the expected reward of each trajectory $\tau$ is a general function $r(\tau)$ which maps any trajectory to a value in $[0, H]$~\citep{zhang24RLHFQ}. Without loss of generality, we scale the average per-step reward into $[0,1]$, and the return of the trajectory does not necessarily need to be the sum of per-step rewards.
For any given policy $\pi$, we can formulate the initial value function $V_1^{\pi}(s)$ as the expected reward of trajectories starting from $s$ with policy $\pi$:
\begin{align*}
    V_1^\pi(s) =& \E_\pi\left[ \left. r(\tau)\right| s_1 = s\right] = \E\left[ \left. r(\tau)\right| s_1 = s, \{a_1,\cdots,a_H\} \sim \pi \right].
\end{align*}
The goal of the RL problem is to find a policy to maximize $V(\pi) = \E_{s\sim \vmu_0}[V_1^\pi(s)]$.

\textbf{Policy Parameterization:} to address the large state space $\sS$ and action space $\sA$ in most RL problems, we assume access to a parameterized policy network $\mathsf{N}_{\vtheta}: \sS \times [H] \to \mathcal{P}(\sA)$ which takes a state and a decision-making step as input, and then outputs the probability distribution of the next action. Here $\vtheta \in \sR^d$ is the policy network parameter vector. Each parameter $\vtheta$ through the policy network will induce a policy which we slightly abuse the notation and use $\pi_{\vtheta}$ to denote.

\textbf{Human Feedback:} The agent has access to human feedback that provides a preference based on the rewards of two trajectories. In each episode, the agent can choose two trajectories $\tau_0$ and $\tau_1$ to query human preference: one-bit feedback $o \in\{0,1\}$. We assume the preference $o$ is generated according to a known preference model where the probability of the outcome between two trajectories is determined by the difference in their rewards. Since the difference is not necessarily a value inside the unit interval, the preference model uses a \emph{link} function $\sigma: \sR \to [0,1]$ to map these differences of rewards to actual probabilities, i.e.,
\begin{align}\label{eq:preference-model}
    \sP(\tau_1 \succ \tau_0) = \sigma(r(\tau_1) - r(\tau_0)),
\end{align}  where $\tau_1\succ\tau_0$ is the event that the human feedback prefers $\tau_1$ over $\tau_0.$
The human feedback $o,$ therefore, is a random sample from a Bernoulli distribution with $\sP(o = 1) = \sP(\tau_1 \succ \tau_0)$.
The notion of link function comes from the dueling bandit literature to model preference with latent utility between arms, e.g. see~\citet[Section 3.2]{Bengs21DuelingBanditSurvey}. This preference model has been used in dueling bandits~\citep{Bengs21DuelingBanditSurvey,Yue09DuelingBandit,Kumagai17DuelingBandit,Ailon14DuelingBandit} as well as RLHF~\citep{Wang23RLHFP2R}. 
One can see that one specific link function $\sigma$ will define a specific preference model~\citep{azari2012randomutility}, i.e., replacing $\sigma(\cdot)$ with a logistic function, we recover the Bradley-Terry model~\citep{Bradley52Preference}, which is commonly used in RLHF for both practical~\citep{Christiano17RLHFTRPO, Ouyang22InstructGPT, Rafailov23DPO} and theoretical works~\citep{Du24RLHFPG,Zhan24RLHFOffline,Zhan24RLHF}. On the other hand, let $\sigma(\cdot)$ be the cumulative distribution function (CDF) of standard normal distribution, we obtain the \emph{Probit} model~\citep{thurstone1927probit} which has wide application in the study of social choice theory, economy, and psychology~\citep{train2009choice,greene2000econometric}. A detailed discussion on other preference models is provided in Section~\ref{sec:relatedwork}.
The following assumption on the link function is standard in both dueling bandits~\citep{Bengs21DuelingBanditSurvey} and preference-based RL~\citep{Wang23RLHFP2R}. One can easily verify that both the Bradley-Terry and Probit models satisfy this assumption.
\begin{assumption}\label{assumpt:link}
    The link function $\sigma(\cdot)$ in the preference model in \eqref{eq:preference-model} is bounded within $[0,1]$ and strictly monotonically increasing on $[-H,H]$ with $\sigma(0) = 1/2$.
\end{assumption}

\textbf{Problem and Notations:} Our goal is to find parameter $\vtheta$ that maximizes the value function, i.e.,
$$
    \max_{\vtheta \in \sR^d} V(\pi_{\vtheta}).
$$
For a scalar $a$, we use $\trim[a|\Delta]$ to represent $\min\{\max\{a, \Delta\}, 1-\Delta\}$. For a vector $\vv$, $\trim[\vv|\Delta]$ represents the vector after applying the $\trim$ operator to each element respectively. Let $\ve_i\in \sR^d$ represent the unit vector with all zero elements but $1$ on the $i$-th coordinate.

\section{Zeroth-Order Policy Gradient Algorithms for RLHF}
In this section, we propose two RLHF algorithms without reward inference, motivated by the relation between preference and zeroth-order gradient. We first present ZPG, a stochastic gradient descent algorithm, and then ZBCPG, a stochastic block-coordinate descent algorithm. 

\subsection{ZPG: Zeroth-Order Policy Gradient from Human Feedback}

\begin{algorithm}[t]
\caption{Zeroth-Order Policy Gradient from Human Feedback}
\label{alg:ZPG}
\Parameters{initial parameter $\vtheta_0$, learning rate $\lr$, trim size $\Delta$, perturbation distance $\mu$.}
\For{$t=1:T$}{
    sample $\vv_t$ uniformly from a unit sphere $\sS^{d-1} = \left\{ \left. \vv \in \sR^d \right| \| \vv \|_2 = 1 \right\}$\;
    obtain a perturbed policy $\pi_{\vtheta_t+\mu \vv_t}$\;
    \For{$n=1:N$}{
        sample trajectory $\tau_{n,0} \sim \pi_{\vtheta_t}$\;
        sample trajectory $\tau_{n,1} \sim \pi_{\vtheta_t + \mu \vv_t}$\;
        query $M$ human evaluators with $(\tau_{n,1}, \tau_{n,0})$ and obtain feedback $[o_{n,1}, \cdots, o_{n,M}]$\;
        estimate preference probability:
        $$
            p_{t,n} = \trim\left[ \left. \sum_{m=1}^M \frac{o_{n,m}}{ M} \right|\Delta\right]\text{\;}
        $$
    }
    estimate the policy gradient:
    $
        \hat{\vg}_t = \frac{d}{ \mu }\frac{\sum_{n=1}^N \sigma^{-1}(p_{t,n})}{N} \vv_t
    $\;
    update $\vtheta_{t+1} = \vtheta_t + \lr \hat{\vg}_t$\; 
}
\end{algorithm}

Our first algorithm ZPG consists of the following five steps in each policy gradient iteration: 
\begin{itemize}
    \item From the current policy $\pi_{\vtheta_t},$ it first obtained a perturbed policy $\pi_{\vtheta_t+\mu \vv_t}$ (line 2-3).

    \item Sample $N$ pairs of trajectories under the two policies $ \pi_{\vtheta_t}$ and $ \pi_{\vtheta_t+\mu\vv_t}$ (lines 5-6). 
    
    \item For each trajectory pair, say $(\tau_{n,1}, \tau_{n,0})$, obtain $M$ independent human preferences (line 7) and estimate the probability that $\tau_{n,1}$ is preferred over $\tau_{n,0}$ (line 8), denoted by $p_{t,n}$.

    \item Use the $N$ estimates $\{p_{t,n}\}_n$ and link function $\sigma(\cdot)$ to estimate the gradient $\hat{\vg}_t$ (line 9).

    \item Update the current policy to a new policy $\vtheta_{t+1}$ using gradient ascent (line 10). 
\end{itemize}

The pseudo-code is presented in  Alg.~\ref{alg:ZPG}. As we mentioned earlier, our approach uses human feedback in a way different from both the classic reward inference in RLHF and DPO. The reward inference uses human preferences to recover the {\em global} reward function and DPO relates the human preference generation mechanism to the optimal policy. We view the human feedback as local information that points to the direction of a more preferred policy, i.e., the policy gradient direction. Some online RLHF algorithms such as online DPO also exploit similar local estimation viewpoints, i.e., using new trajectories of current policy to locally improve the estimation of DPO loss and then proceed. However, we want to emphasize that the relation between the DPO loss and the optimal policy is still global, and it is limited to deterministic MDPs, which shows the novelty of our approach. The algorithm we propose has two key components: (i) a value function difference estimator from human preference and (ii) a policy gradient estimator from value function difference.

\textbf{Policy Gradient Approximation:} At each iteration of the algorithm, it first samples a $d$ dimensional vector $\vv_t$ from a unit sphere and then perturbs the policy network parameter $\vtheta_t$ along the direction of this sampled vector. Then, it uses the inner for loop to construct an estimation of the value function difference between the original policy and the perturbed policy, i.e., $V(\pi_{\vtheta_t + \mu \vv_t}) - V(\pi_{\vtheta_t})$. We then plug it into the zeroth-order stochastic gradient descent (SGD) algorithm proposed in~\citep{Ghadimi13ZOSGD} to construct a zeroth-order approximation to the policy gradient, i.e.,
\begin{align*}
    \nabla_{\vtheta} V(\pi_{\vtheta_t}) \approx \E_{\vv_t}\left[ \frac{d}{\mu} \left( V(\pi_{\vtheta_t + \mu \vv_t}) - V(\pi_{\vtheta_t}) \right) \vv_t \right].
\end{align*}
We remark that the random vector $\vv_t$ for each iteration can also be drawn from a normal distribution~\citep{nesterov17zosgd}, but the unit sphere is more numerically stable.

\textbf{Value Function Inference:} The inner for loop of Alg.~\ref{alg:ZPG} aims to estimate the value function difference between the perturbed policy $\pi_{\vtheta_t + \mu \vv_t}$ and current policy $\pi_{\vtheta_t}$. The algorithm samples multiple trajectory pairs with both policies and for each pair, it queries humans multiple times to obtain pairwise preferences $[o_{n,1}, \cdots, o_{n,M}]$. It then uses the preferences to construct a robust estimator $p_{t,n}$ to approximate the probability of comparison $\sP(\tau_{n,1} \succ \tau_{n,0})$, which is further converted to an estimate of the value function difference based on the preference model in~\eqref{eq:preference-model} as follows:
\begin{align}\label{eq:preference-estimate}
    V(\pi_{\vtheta_t + \mu \vv_t}) - V(\pi_{\vtheta_t}) \approx \frac{1}{N}\sum_{n=1}^N \sigma^{-1}(p_{t,n}).
\end{align}
Querying humans multiple times ensures an accurate estimation of the reward gap between two trajectories. The reward gap of two trajectories is a random sample of the value function difference, so we sample multiple trajectories to ensure the average trajectory reward gap converges to the value function difference. To ensure finite variance after applying $\sigma^{-1}(\cdot)$ function, we trim $p_{t,n}$ with a small constant which can be set to $\min\{\sigma(-H), 1 - \sigma(H)\}$ in this case.

\textbf{Remarks on Local and Global Reward Estimation.} The main philosophical difference between our proposed algorithms and classic RLHF paradigms is the use of local reward information around the neighborhood of the current policy. This is completely different from RLHF with reward inference since reward models intend to approximate the global reward landscape of the MDP. Some recently developed online RLHF algorithms, such as online DPO~\citep{guo2024onlineDPO} and XPO~\citep{xie2024xpo} also employ certain local reward information to estimate the DPO loss in each iteration. Specifically, the algorithms will explore the trajectories in the neighborhood of the current policy to construct a local estimation of the DPO loss. However, on the other hand, the relation between the DPO loss and the best policy is still global in deterministic MDPs studied in both papers. Our paper's approach is purely local and does not rely on such global relations, which is incorrect in stochastic MDPs. It estimates the local policy gradient directly from human feedback, and proceeds gradient descent for policy updates. The use of local information in our approach is more general and may apply to a wider range of RL problems.

\subsection{ZBCPG: Zeroth-Order Block-Coordinate Policy Gradient from Human Feedback}

\begin{algorithm}[t]
\caption{ Zeroth-Order Block-Coordinate Policy Gradient from Human Feedback}
\label{alg:ZBCPG}
\Parameters{initial parameter $\vtheta_0$, learning rate $\lr$, trim size $\Delta$, perturbation distance $\mu$, coordinate batch size $K$.}
\For{$t=1:T$}{
    sample a set of $K$ coordinates $\vi_t = [i_{t,1}, i_{t,2}, \cdots, i_{t,K}]$ from $\{1,2,\cdots, d\}$\;
    sample a set $\boldsymbol{\lambda}_t = [\lambda_{t,1}, \lambda_{t,2}, \cdots, \lambda_{t,K}]$ where each $\lambda_{t,j}$ is uniformly sampled from $\{-1,1\}$\;
    construct the perturbation vector:
    $
        \vv_t = \frac{1}{\sqrt{K}}\sum_{j=1}^K \lambda_{t,j} \ve_{i_{t,j}}
    $\;
    \For{$n=1:N$}{
        sample trajectory $\tau_{n,0} \sim \pi_{\vtheta_t}$\;
        sample trajectory $\tau_{n,1} \sim \pi_{\vtheta_t + \mu \vv_t}$\;
        query $M$ human evaluators with $(\tau_{n,1}, \tau_{n,0})$ and obtain feedback $[o_{n,1}, \cdots, o_{n,M}]$\;
        estimate preference probability:
        $$
            p_{t,n} = \trim\left[ \left. \sum_{m=1}^M \frac{o_{n,m}}{ M} \right|\Delta\right];
        $$
    }
    estimate the policy gradient:
    $
        \hat{\vg}_t = \frac{d}{ \mu }\frac{\sum_{n=1}^N \sigma^{-1}(p_{t,n})}{N} \vv_t
    $\;
    update $\vtheta_{t+1} = \vtheta_t + \lr \hat{\vg}_t$\; 
}
\end{algorithm}

In high-dimensional optimization problems, it is usually memory restricted or operation inefficient to approximate the full gradient and update all the parameters in the policy network at the same iteration step~\citep{Malladi23ZOLLM,zhang2024ZOLLM}, which motivates parameter efficient fine-tuning (PEFT). The stochastic (block) coordinate descent approach naturally arises because of its ease of implementation, low memory requirements, and adaptability to distributed settings~\citep{nesterov12scd,lu15bscd}. The same advantage also applies to RLHF when the number of parameters in the policy network is too large. Therefore, we propose ZBCPG, a block coordinate version of ZPG, which is summarized in Alg.~\ref{alg:ZBCPG}. The key difference between ZBCPG and ZPG is the choice of the perturbation direction, where we use Rademacher noise instead of the normal perturbation in ZPG. 

\textbf{Zeroth-Order Block Coordinate Gradient Approximation:}  Instead of sampling from a sphere, which perturbs all parameters of the policy network,  ZBCPG separates the sampling procedure into two simple parts: first sample a minibatch of coordinates and then sample a zero-centered Bernoulli random variable for each coordinate, which still results in a valid gradient estimator. 
\begin{align*}
    \nabla_{\vtheta} V(\pi_{\vtheta_t}) \approx \E_{\vi_t, \boldsymbol{\lambda}_t}\left[ \frac{d}{K\mu} \left( V(\pi_{\vtheta_t + \mu \vv_t}) - V(\pi_{\vtheta_t}) \right) \vv_t \right].
\end{align*}
The block-coordinate approach allows us to (i) perturb a subset of parameters at each iteration, e.g., a specific layer of the policy network for fine-tuning, and (ii) have a parallel implementation where we have multiple gradient estimators $\hat{\vg}_t$ when updating the policy. We will later show that both algorithms have similar provable convergence guarantees but the analysis of ZBCPG is more challenging due to the perturbation mechanism.

\textbf{Remarks on KL Regularization in RLHF:} the KL-regularization is commonly used in RLHF for both empirical and theoretical studies, e.g., PPO~\citep{Ouyang22InstructGPT}, DPO~\citep{rafailov24dpo,rafailov24dpoQ,guo2024onlineDPO,dong24onlineDPO}, and more recent algorithms~\citep{xie2024xpo,xiong24iterativeDPO}. On the positive side, the KL-regularization improves the training stability in RLHF tasks. For example, in LLM applications, the model pre-training procedure will provide powerful capabilities such as reasoning, and the regularization ensures that the updated policy is unlikely to deviate from the pre-training model and lose these capabilities. It also ensures the global relation between the optimal policy and the reward function in deterministic MDPs for LLM studies and motivates easier objectives such as the DPO loss. On the other hand, in general RL problems, if the regularization is too large, it also limits the RL agent from the possibility of learning the best policy, which may be very different from the reference policy of the pre-trained model. It is also worth noting that KL-regularized MDPs may be easier to explore compared to their unregularized counterparts, especially for MDPs with sparse rewards. This is because the best policies for regularized MDPs are likely to be stochastic policies which exhibit certain exploration power due to stochasticity. In contrast, the optimal policy for unregularized counterparts is deterministic, which does not explore. It seems there is no golden rule on whether KL-regularization is indeed needed in RLHF applications beyond LLMs. The answer is likely to depend on specific applications. Our algorithm can also be applied to KL-regularized MDPs after a slight modification in gradient estimation. In this case, the zeroth-order gradient of the policy will include the sum of two components, the value function difference estimation, and the KL regularization difference, i.e.,
\begin{align*}
    \left(V(\pi_{\vtheta_t + \mu \vv_t}) - \beta \mathsf{KL}(\pi_{\vtheta_t + \mu \vv_t}\| \pi_{\vtheta_0})\right) -   \left(V(\pi_{\vtheta_t}) - \beta \mathsf{KL}(\pi_{\vtheta_t}\| \pi_{\vtheta_0})\right),
\end{align*}
since the objective is now the KL-regularized value function. Here, $\pi_{\vtheta_0}$ is the reference policy, and $\mathsf{KL}(\pi_\vtheta\| \pi_{\vtheta_0})$ is the KL divergence between a policy parameterized by $\vtheta$ and the reference policy, i.e.,
\begin{align*}
    \mathsf{KL}(\pi_\vtheta\| \pi_{\vtheta_0}) = \E_{s_h, a_h\sim \pi_\vtheta}\left[ \sum_{h=1}^H \log \frac{\pi_\vtheta(a_h|s_h)}{\pi_{\vtheta_0}(a_h|s_h)}\right].
\end{align*}
Remark that human evaluators will only provide preference based on the trajectory returns, not the KL regularization. Therefore, the value function difference component can be estimated the same way as the current algorithm, i.e., line $9$ in Alg.~\ref{alg:ZPG}. Then, the rest is to estimate the KL regularization difference between the perturbed policy $\pi_{\vtheta_t + \mu \vv_t}$ and the current policy $\pi_{\vtheta_t}$, concerning the reference policy $\pi_{\vtheta_0}$, i.e.,
\begin{align*}
    \mathsf{KL}(\pi_{\vtheta_t + \mu \vv_t}\| \pi_{\vtheta_0}) -& \mathsf{KL}(\pi_{\vtheta_t}\| \pi_{\vtheta_0})\\ 
    =& \E_{s_h, a_h\sim \pi_{\vtheta_t + \mu \vv_t}}\left[ \sum_{h=1}^H \log \frac{\pi_{\vtheta_t + \mu \vv_t}(a_h|s_h)}{\pi_{\vtheta_0}(a_h|s_h)}\right] - \E_{s_h, a_h\sim \pi_{\vtheta_t}}\left[ \sum_{h=1}^H \log \frac{\pi_{\vtheta_t}(a_h|s_h)}{\pi_{\vtheta_0}(a_h|s_h)}\right].
\end{align*}
This can be achieved by estimating the KL regularization term of both policies, which is conducted by evaluating the action probability of the reference policy $\pi_{\vtheta_0}$ on the trajectories generated from both policies. With both components, we can construct the zeroth-order gradient for the regularized MDP and conduct gradient descent similar to ZPG and ZBCPG.

\section{Theoretical Analysis: Rate of Convergence}
In this section, we provide theoretical performance guarantees for both ZPG and ZBCPG. We first provide technical assumptions on the preference generation model, the policy network, and the value function landscape, which is necessary for deriving theoretical insights.

\subsection{Assumptions}
To infer the local reward difference from human preference probability through link function $\sigma(\cdot)$, we impose the following assumption which is satisfied by the Bradley-Terry model. A slightly weaker assumption is also adopted by~\citet{Wang23RLHFP2R} and justified as a minimal requirement to learn the optimal policy. We use $\Delta = \min\{\sigma(-H), 1-\sigma(H)\}$ as the trim constant.
\begin{assumption}\label{assumpt:linkgrad}
     The inverse link function $\sigma^{-1}(\cdot)$ is $L$-Lipchitz continuous on $[\Delta, 1-\Delta]$.
\end{assumption}
We further require the landscape of the value function and the policy network to be ``regular'', and impose the following assumption, which is a standard assumption used in nonconvex optimization literature~\citep{liu18zosignsgd,bernstein18signsgd,reddi18adam}.
\begin{assumption}\label{assumpt:smooth}
    The value function $V(\pi_{\vtheta})$ for the policy network parameters $\vtheta$ is $L$-smooth on $\sR^d$.
\end{assumption}
Since a trajectory reward is bounded in $[0, H]$,  $V(\pi_{\vtheta^*})<\infty$, where $\vtheta^*$ is the global optimal solution. For simplicity, we assume $L$ is the constant upper bound for both assumptions.

\subsection{Convergence Rate and Sample Complexity}
In this section, we present the theoretical guarantees for both ZPG and ZBCPG under all three assumptions mentioned in previous sections. We aim to learn an $\eps$-stationary policy $\pi_{\vtheta}$ with $\|\nabla_{\vtheta} V(\pi_\vtheta)\|_2^2 \leq \eps$, and study the convergence rate and sample complexity.  
\begin{theorem}\label{thm:ZPG}
    Choose the perturbation distance $\mu$ and the learning rate $\alpha$ to be chosen as follows:
    \begin{align*}
        \mu ^2 = \Theta\left(\max\left\{\frac{1}{\sqrt{M}}, \frac{H}{\sqrt{d N}} \right\}\right), \quad \alpha = \Theta\left(\frac{1}{d}\right).
    \end{align*}
    If $M = \Omega(H^2)$ and we randomly pick $\vtheta_R$ uniformly from the trajectory $\{\vtheta_0, \vtheta_1, \cdots, \vtheta_{T-1}\}$, then the convergence rate of ZPG satisfies:
    \begin{align*}
        \E\left[\|\nabla_{\vtheta} V(\pi_{\vtheta_R})\|_2^2\right] = \gO\left( \frac{H d}{T} + \frac{d^2 \sqrt{\log M}}{\sqrt{M}} + \frac{H d\sqrt{d}}{\sqrt{N}} \right).
    \end{align*}
\end{theorem}
\begin{theorem}\label{thm:ZBCPG}
    Choose the perturbation distance $\mu$ and the learning rate $\alpha$ to be chosen as follows:
    \begin{align*}
        \mu ^2 = \Theta\left(\max\left\{\frac{1}{\sqrt{M}}, \frac{H}{\sqrt{d N}} \right\}\right), \quad \alpha = \Theta\left(\frac{1}{d}\right).
    \end{align*}
    If $M = \Omega(H^2)$ and we randomly pick $\vtheta_R$ uniformly from the trajectory $\{\vtheta_0, \vtheta_1, \cdots, \vtheta_{T-1}\}$, then the convergence rate of ZBCPG satisfies:
    \begin{align*}
        \E\left[\|\nabla_{\vtheta} V(\pi_{\vtheta_R})\|_2^2\right] = \gO\left( \frac{H d}{T} + \frac{d^2 \sqrt{\log M}}{\sqrt{M}} + \frac{H d\sqrt{d}}{\sqrt{N}} \right).
    \end{align*}
\end{theorem}
The complete proof of both theorems is presented in the appendix. Here we first provide insights into the choice of hyper-parameters and convergence rate results in both theorems, and then we discuss the challenges and technical novelties of our proof.

\textbf{Insights behind the Convergence Rate:} Both ZPG and ZBCPG have the same rate of convergence which consists of three components: the zeroth-order gradient descent rate, the preference estimation error, and the value function approximation error
\begin{align*}
    \underbrace{\frac{Hd}{T}}_{\text{Zeroth-Order Gradient Descent}} + \underbrace{\frac{d^2\sqrt{\log M}}{\sqrt{M}}}_{\text{Preference Estimation}} + \underbrace{\frac{H d\sqrt{d}}{\sqrt{N}}.}_{\text{Value Function Approximation}}
\end{align*}
The second represents the error that occurs when using multiple human preferences $[o_{n,1}, \cdots, o_{n,M}]$ to approximate the population-level human preference probability for given two trajectories, i.e., $\sP(\tau_{n,1} \succ \tau_{n,0})$. This error will further result in a bias term after being plugged into the inverse link function $\sigma^{-1}(\cdot)$ to construct an estimation of the value function difference. The third term comes from the variance of using multiple trajectory rewards to approximate the value function of a policy. The first term represents the error resulting from zeroth-order stochastic gradient descent or blocked coordinate descent, which matches the state-of-the-art analysis result $\gO(d/T)$ for non-convex smooth function optimization~\citep{nesterov17zosgd}. Even though the final convergence rates are the same and we both use constant learning rates, how we choose the perturbation distance to obtain the rate differs from~\citep{Ghadimi13ZOSGD}. Specifically, they chose a small perturbation distance with $\mu^2 = \gO(d/T)$ to make sure the zeroth order approximation error is of lower order. However, this choice will not work for us, because our gradient estimate is biased due to the non-linear nature of the link function in preference estimation. If we choose the perturbation distance $\mu$ to be too small, the preference estimation error will be amplified by $d/\mu$ due to the formula of zeroth-order approximation $\hat{\vg}_t$. This phenomenon adds complication to our theoretical analysis. Our method is to use a moderate perturbation distance $\mu$. Moreover, this moderate perturbation distance also balances the preference estimation and the value function approximation errors. 

Based on the theorems, we have the following corollary that characterizes the sample complexity.
\begin{corollary}
To learn an $\eps$-stationary policy, the required number of human preference queries of ZPG and ZBCPG with proper hyper-parameters satisfies
    \begin{align*}
        TMN = \gO\left( \frac{d^8 H^3}{\eps^5} \log \left( \frac{d}{\eps} \right) \right).
    \end{align*}
\end{corollary}

\textbf{Remarks on Local and Global Convergence.} Our results only establish the local convergence of both proposed algorithms. In general, considering the generality of stochastic MDPs and policy parameterization studied in our paper, it is extremely difficult to establish global convergence results without further assumptions beyond the ones that have already been made in the paper, such as smoothness. We conjure that our proposed algorithms will have global convergence under additional assumptions on the value function, such as convexity or the Polyak-Lojasiewicz (PL) condition~\citep{karimi2016PL}. Again, these assumptions are often questionable in practice. Some contemporary works, such as~\citet{xie2024xpo}, established global convergence in the deterministic KL-constrained MDPs with additional coverability assumptions, a much easier and narrowed model where the optimal policy can be explicitly expressed.
Considering the difficulty and generality of our setting, we believe local convergence is the best result that can be obtained, and a meaningful result showing the effectiveness of the proposed algorithm to a wider range of RL problems beyond LLMs. It remains an open question of how to design algorithms that achieve global optimality under minimal and practical assumptions.

\subsection{Technical Challenges and Proof Novelties}\label{sec:proofnovelty}
In this section, we first overview the proof of zeroth-order stochastic gradient descent used in~\citep{Ghadimi13ZOSGD,nesterov17zosgd,Gao18ZOSGDsmoothing,liu18zosignsgd}
from a Lyapunov drift optimization perspective. We then show the major technical difficulties in applying such a framework to analyze both ZPG and ZBCPG, i.e., the gradient estimator is biased due to stochastic human preference. Then, we demonstrate our novel analysis techniques to resolve them.

\textbf{Classic Proof of Zeroth-Order Optimization:} to illustrate the procedure of the analysis of zeroth order gradient estimate, we suppose we can query $V(\pi_{\vtheta})$ for any $\vtheta$. This procedure makes use of the randomized smoothing function $V_\mu(\vtheta)$~\citep{Ghadimi13ZOSGD,Gao18ZOSGDsmoothing} as
$$
    V_\mu(\pi_{\vtheta}) = \E_{\vv'}\left[ V(\pi_{\vtheta + \mu \vv'}) \right],
$$
where the random vector $\vv'$ follows a uniform distribution over the unit Euclidean ball. It is shown in~\citep{Gao18ZOSGDsmoothing} that the zeroth-order gradient estimator used in ZPG, constructed from sampling $\vv_t$ uniformly over a sphere, is an unbiased estimator of the smoothing function gradient, i.e.,
\begin{align*}
    \nabla_{\vtheta} V_\mu(\pi_{\vtheta}) =\E_{\vv}\left[ \frac{d}{\mu} \left( V(\pi_{\vtheta + \mu \vv}) - V(\pi_{\vtheta_t}) \right) \vv \right], 
\end{align*}
where $\vv$ is sampled from a unit sphere. Clearly, the gradient of the smoothing function is not equal to the original value function, as well as the function itself, but it can be shown that they are close as long as $\mu$ is small~\citep{liu18ZOSVRG}:
\begin{align}\label{eq:zoapprox}
    |V_\mu(\pi_{\vtheta}) - V(\pi_{\vtheta})| = \gO\left( \mu^2\right); \quad 
    \|\nabla_{\vtheta} V_\mu(\pi_{\vtheta}) - \nabla_{\vtheta} V(\pi_{\vtheta})\|_2 = \gO\left( \mu d\right).
\end{align}
The standard proof uses the randomized smoothing function $V_\mu(\pi_{\vtheta})$ as the Lyapunov function and then bounds the drift given the stochastic gradient descent update rule when $\alpha = \Theta( 1/d)$. Neglecting problem-independent constants, we have:
\begin{align*}
    V_\mu(\pi_{\vtheta_t}) - &V_\mu(\pi_{\vtheta_{t+1}})\\ 
    \leq & - \alpha \underbrace{\|\nabla_{\vtheta} V_\mu(\pi_{\vtheta_t})\|_2^2}_{\mathsf{Drift}} + \alpha \underbrace{\langle \nabla_{\vtheta} V_\mu(\pi_{\vtheta_t}), \nabla_{\vtheta} V_\mu(\pi_{\vtheta_t}) - \hat{\vg}_t }_{\text{$1$st Order: }\mathsf{GradBias}} \rangle + \alpha^2\underbrace{\|\hat{\vg}_t - \nabla_{\vtheta} V_\mu(\pi_{\vtheta_t})\|_2^2}_{\text{2nd Order: }\mathsf{GradVar}\approx \mu^2 d^2}.
\end{align*}
Note the gradient estimator $\hat{\vg}_t$ is unbiased and bounded, and the gradient of $V_\mu(\pi_{\vtheta})$ is close to $V(\pi_{\vtheta})$, taking a conditional expectation over the filtration before time $t$ will result in:
\begin{align*}
    \E[V_\mu(\pi_{\vtheta_t})| \gF_t] - &\E[V_\mu(\pi_{\vtheta_{t+1}})| \gF_t] \\
    \leq & - \alpha \|\nabla_{\vtheta} V_\mu(\pi_{\vtheta_t})\|_2^2 + \alpha \langle \nabla_{\vtheta} V_\mu(\pi_{\vtheta_t}), \E[\nabla_{\vtheta} V_\mu(\pi_{\vtheta_t}) - \hat{\vg}_t] \rangle + \alpha^2 \mu^2 d^2\\
    \leq &  - \alpha \|\nabla_{\vtheta} V(\pi_{\vtheta_t})\|_2^2 + \alpha \mu^2 d^2 + \alpha^2 \mu^2 d^2,
\end{align*}
where the last step uses~\eqref{eq:zoapprox} and the fact that the gradient is unbiased. Let us choose a small learning rate $\alpha = \Theta(1/d)$ and take an expectation with a telescoping sum to obtain:
\begin{align*}
    \underbrace{\frac{\E[V_\mu(\pi_{\vtheta_0})] - \E[V_\mu(\pi_{\vtheta_T})]}{T}}_{\gO(H/T)} \lesssim -\alpha \underbrace{ \E\left[\frac{\sum_{t=1}^T \|\nabla_{\vtheta} V(\pi_{\vtheta_t})\|_2^2}{T} \right]}_{\mathsf{Target}} + \alpha \mu^2 d^2.
\end{align*}
A little manipulation will lead to the following bound, which can be made small when $\mu \approx \sqrt{1/dT}$.
\begin{align*}
    \mathsf{Target} = \gO\left(\frac{H}{T\alpha} + \mu^2 d^2 \right)  = \gO\left(\frac{H d}{T} + \mu^2 d^2\right) = \gO\left( \frac{H d}{T} \right).
\end{align*}

\textbf{Amplified Gradient Biases for ZPG:} If we directly apply the steps above to ZPG, we immediately run into the issue that our gradient estimator $\hat{\vg}_t$ in expectation is biased even compared to smoothing function gradient due to preference estimation. Moreover, the second-order gradient variance will be larger since we used trajectory reward to estimate the value function. From concentration, we will obtain an error bound of using preference to estimate the value function difference as:
\begin{align*}
    \left| \E\left[\sigma^{-1}(p_{t,n})\right] -  \left( V(\pi_{\vtheta_t + \mu \vv_t}) - V(\pi_{\vtheta_t}) \right) \right| \leq \tilde{\gO}\left( \frac{1}{\sqrt{M}} \right),
\end{align*}
where $\tilde{\gO}$ hides logarithmic terms. This bias term will be amplified by $d/\mu$ and then added to the gradient estimation bias in the first-order drift term if plugged into the analysis:
\begin{align*}
    \E[V_\mu(\pi_{\vtheta_t})| \gF_t] - \E[V_\mu(\pi_{\vtheta_{t+1}})| \gF_t] \leq - \underbrace{\alpha \|\nabla_{\vtheta} V(\pi_{\vtheta_t})\|_2^2 + \alpha \mu^2 d^2}_{\text{Same Drift as Before}} + \alpha \underbrace{\frac{d \|\nabla_{\vtheta} V_\mu(\pi_{\vtheta_t})\|_2}{\mu \sqrt{M}}}_{\text{Additional Bias}}.
\end{align*}
Using the same perturbation distance as before, the additional bias will lead to an $\tilde{\gO}(\sqrt{T/M})$ term in the final bound, which is small only when $M$ is much larger than $T$ and is much looser compared with ours. For example, letting $M=T^2,$ the above bound is $\tilde{\gO}(1/\sqrt{T})$ while ours is $\tilde{\gO}(1/T).$

Our approach to avoid this term in the final result is to make use of the gradient value $\nabla_{\vtheta} V(\pi_{\vtheta_t})$ in the first-order term to cancel out the additional bias on certain occasions. Specifically, we divide the trajectory of $\vtheta_t$ into two sets, one with a relatively large gradient and one with a relatively small gradient. For $\vtheta_t$ with a large gradient, we use a part of the negative drift to cancel out the additional bias, since the negative drift is the square of the gradient $\nabla_{\vtheta} V(\pi_{\vtheta_t})$ which is even larger. For $\vtheta_t$ with a small gradient, we know the bias term will be small and thus can provide a refined drift bound. Combining this analysis with a slightly larger perturbation distance $\mu$, we will be able to balance the additional bias with gradient variance to cancel out the $\tilde{\gO}(\sqrt{T/M})$ term and obtain the final result.

\textbf{Implicit Smoothing Function for ZBCPG:} Due to the choice of blocked perturbation vector sampling procedure, it is difficult to obtain the exact analytical expression of the smoothing function $V_\mu(\pi_{\vtheta})$ whose gradient is the expectation of gradient estimation $\hat{\vg}_t$ for ZBCPG. This prohibits us from continuing to use $V_\mu(\pi_{\vtheta})$ as the Lyapunov function, as it is hard to analyze the gradient bias and the variance without an explicit target format. However, if we rethink the reason for introducing the smoothed function in zeroth-order optimization, we hope the gradient of the smoothed function will be unbiased to cancel out the first-order positive drift. However, this is already not true in the analysis of ZPG since we have gradient estimation bias from human feedback, but it is small enough on average to be controlled. If the gradient difference between the smoothed function $V_\mu(\pi_{\vtheta})$ and the vanilla value function $V(\pi_{\vtheta})$ is smaller than this additional bias, then we can use the original value function $V(\pi_{\vtheta})$ as the Lyapunov function at the cost of an additional bias besides preference estimation. 
Fortunately, this can be achieved through a carefully chosen perturbation distance $\mu$ to balance these two types of errors.

\section{Experiments}\label{sec:experiment}

In this section, we study the empirical performance of ZPG and ZBCPG in a stochastic GridWorld environment to show the practical value of both algorithms, which coincides with the theoretical results. We consider three baselines: (1) RM+PPO~\citep{Ouyang22InstructGPT}, (2) DPO for token-level MDP~\citep{rafailov24dpoQ}, and (3) Online DPO for token-level MDP~\citep{dong24onlineDPO,guo2024onlineDPO}, where the implementation detail can be found in the appendix.

\textbf{Environment.} Our GridWorld environment consists of $5 \times 5$ blocks, where the location of each block, i.e., the state space $\sS$, is denoted as $(1,1)$ to $(5,5)$. For each block, we flip a fair coin. If the head shows, we will place a random reward on top, and if the tail shows, there will be no reward on this block. The reward, if placed in a block, is randomly sampled from a standard normal distribution, which means we allow a negative reward as punishment. Each episode consists of $H=20$ steps, and at the start of each episode, an agent is positioned in block $(3,3)$, i.e., the center of the GridWorld environment. At each step, the agent can choose from four actions that constitute the action space $\sA$: going up, going down, going left, and going right. However, unlike classic GridWorld environments where the agent can perfectly control its movement, in our experiment, we assume the agent is subject to environment disturbances or imperfect control such as wind or turbulence, to model the stochastic transition and demonstrate the weaknesses of predominant RLHF frameworks such as DPO. We assume at each step, the action taken by the agent may be reversed with probability $0.4$ due to disturbances. For example, if the agent is in the center and chooses to go up, it will land on the block above the current block with a probability of $0.6$ and on the block beneath it with a probability of $0.4$. Similarly, if the agent chooses to go left, it will land on the block left next to it with a probability of $0.6$ and on the block right next to it with a probability of $0.4$. The motivation for imperfect control arises naturally from designing agents for turbulent environments, e.g., for robots operating on airplanes, cars, or ships, where continual wind or bump may shift the agent's action. The goal of the agent is to maximize the cumulative reward over the $H=20$ time horizon, and the interaction is conducted episodically.

\textbf{Policy Parameterization.} For all algorithms implemented in our experiment, we used tabular policy softmax parameterization, i.e., each state-action pair $(s,a)$ is equipped with a parameter $\xi_{s,a}$ and the policy $\pi(a|s)$ of taking action $a$ at state $s$ would follow:
\begin{align*}
    \pi(a|s) = \frac{\exp(\xi_{s,a})}{\sum_a \exp(\xi_{s,a})}.
\end{align*}

\textbf{Human Feedback.} In this experiment, we assume access to a human panel of size $M=1000$, i.e., a group of human experts, can be recruited and queried at the same time for two trajectories, except for the distributed implementation of ZBCPG, which we will discuss in more detail later. Each human expert, which we refer to as a panelist, will generate a preference based on a specific preference model characterized by a link function. The agent can query human evaluators for preference over two trajectories. We consider two preference models: (1) Bradley-Terry model with a logistic link function and (2) Weibull model with an anti-asymmetric Weibull link function, both over the trajectory rewards. For example in Bradley-Terry model, for two trajectories $\tau_0$ and $\tau_1$, each panelist would provide feedback following the distribution as follows:
\begin{align*}
    \mathbb{P}(\tau_1\succ \tau_0) = \frac{\exp(r(\tau_1))}{\exp(r(\tau_1)) + \exp(r(\tau_0))}, \quad \mathbb{P}(\tau_0\succ \tau_1) = \frac{\exp(r(\tau_0))}{\exp(r(\tau_1)) + \exp(r(\tau_0))}.
\end{align*}
For the Weibull preference feedback mechanism~\citep{train2009choice}, it is worth noting that the link function is anti-asymmetric between the two trajectories being compared, i.e., $\sigma(x) \neq 1 - \sigma(-x)$. For example, if two trajectories $(\tau_0, \tau_1)$ are compared by a panelist with the Weibull preference model, it would prefer the trajectory $\tau_1$ with probability:
\begin{align*}
    \mathbb{P}(\tau_1 \succ \tau_0) = \exp_2(- \exp_2(r(\tau_0) - r(\tau_1))).
\end{align*}
This preference model would capture the bias effect presented in the revealing sequence to the human evaluators. Namely, human experts would unintentionally find it easier to distinguish if the better trajectory is presented first.

\subsection{Results on Bradley-Terry Human Feedback}
\begin{figure}[ht]
    \centering
    \subfigure[performance]{
        \centering
         \includegraphics[width=0.48\linewidth]{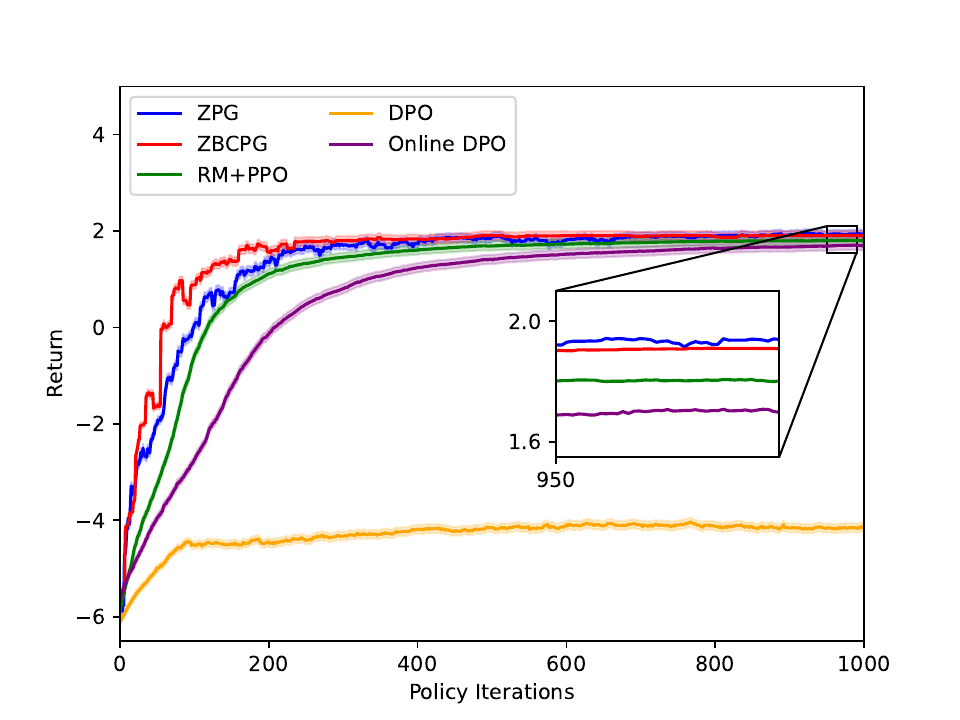}
         }
     \subfigure[parallelization]{
        \centering
         \includegraphics[width=0.48\linewidth]{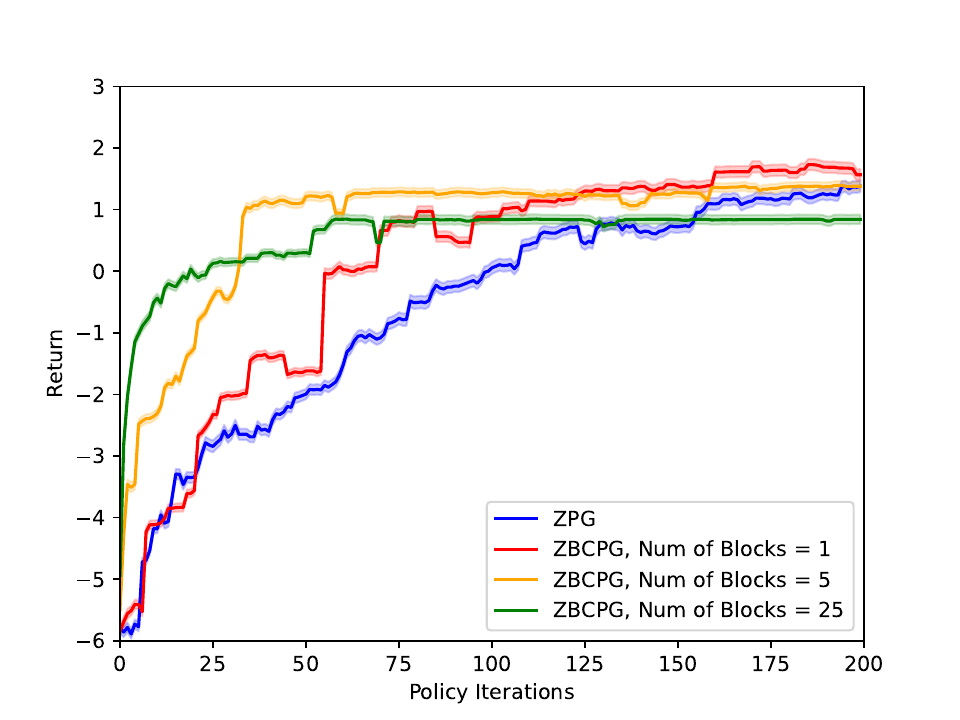}
         }
    \caption{GridWorld with Bradley-Terry Feedback: (a) the trajectory return of ZPG, ZBCPG, and RLHF baselines, and (b) the return of ZBCPG with different parallelization levels. All results are averaged over $10^5$ repetitions of policy evaluation and shaded areas indicate confidence intervals.}
    \label{fig:reward}
\end{figure}
\begin{table}[ht]
    \centering
    \begin{tabular}{cccccc}
    \toprule
    Algorithm     & ZPG ({\color{blue} Ours}) & ZBCPG ({\color{blue} Ours}) & RM+PPO & DPO & Online DPO \\
    \midrule
    Return    &$\boldsymbol{1.94} \pm 0.09$ & $\boldsymbol{1.91} \pm 0.09$ & $1.80\pm 0.09$ & $-4.13 \pm 0.09$ & $1.71 \pm 0.09$ \\
    \bottomrule
    \end{tabular}
    \caption{Last Iterate Policy Average Return with Bradley-Terry Feedback.}
    \label{tab:return}
\end{table}

\textbf{Comparison of Average Return.} We first study and compare the empirical performance of all algorithms with Bradley-Terry human feedback. All algorithms, including both our proposed ZPG and ZBCPG with all baselines, collect $N=1000$ trajectory pairs between policy updates and $M=1000$ human experts evaluate each pair. The trajectory return for each iteration is compared in Fig.~\ref{fig:reward}(a) and the return of the final policy is reported in Tab.~\ref{tab:return}. Both ZPG and ZBCPG perform better than the three baselines in both convergence rates and the quality of the last iterate policy. Compared to PPO, our algorithms converge to a better policy partially since the reward model is inaccurate and the agent is not able to learn the optimal policy from it. It is also observed that vanilla DPO has a much worse performance than our proposed algorithms. This may result from two reasons: first, the DPO loss is valid only in deterministic MDP, and second, DPO is constrained to the neighborhood of the sub-optimal reference policy. The online DPO algorithm improves over vanilla DPO but still has inferior performance due to the inherent model error of the DPO loss. This also shows the fundamental difference between stochastic and deterministic MDPs, and the need to design RLHF algorithms for general RL problems. Moreover, online DPO converges much slower, partly because the DPO loss landscape becomes flat and hard to optimize when the weight of the KL constraint is small for better exploration. 

\textbf{Power of Parallelization.}
We also compare our proposed algorithms to distributed implementations of ZBCPG to show the influence of parallelization of the block-coordinate descent. 
For distributed ZBCPG, we assume multiple parallel agents can collect trajectories at the same time and query humans at the same time. However, the panel of human evaluators is also separated into small groups for parallelization and each agent can only work with a small group of experts to perform optimization. 
The results are shown in Fig.~\ref{fig:reward}(b) where the number of blocks shown in the legend controls the level of parallelization. A detailed description of the implementation is presented in the appendix. It is shown that as the number of blocks increases, ZBCPG converges faster to a stationary policy. However, the number of human queries per pair of trajectories in each parallelization also decreases, which introduces a larger gradient bias and leads to a sub-optimal policy. Therefore, the trade-off between computation parallelization and accuracy should be taken carefully.

\subsection{Results on Weibull Human Feedback}
In this section, we conduct additional numerical experiments on Weibull human feedback to evaluate the impact of the preference model link functions on our proposed algorithms and the baselines, as well as the influence of the size $M$ of the human evaluator panel. 
\begin{figure}[t]
    \centering
    \subfigure[Weibull, $M=1000$]{
        \centering
         \includegraphics[width=0.48\linewidth]{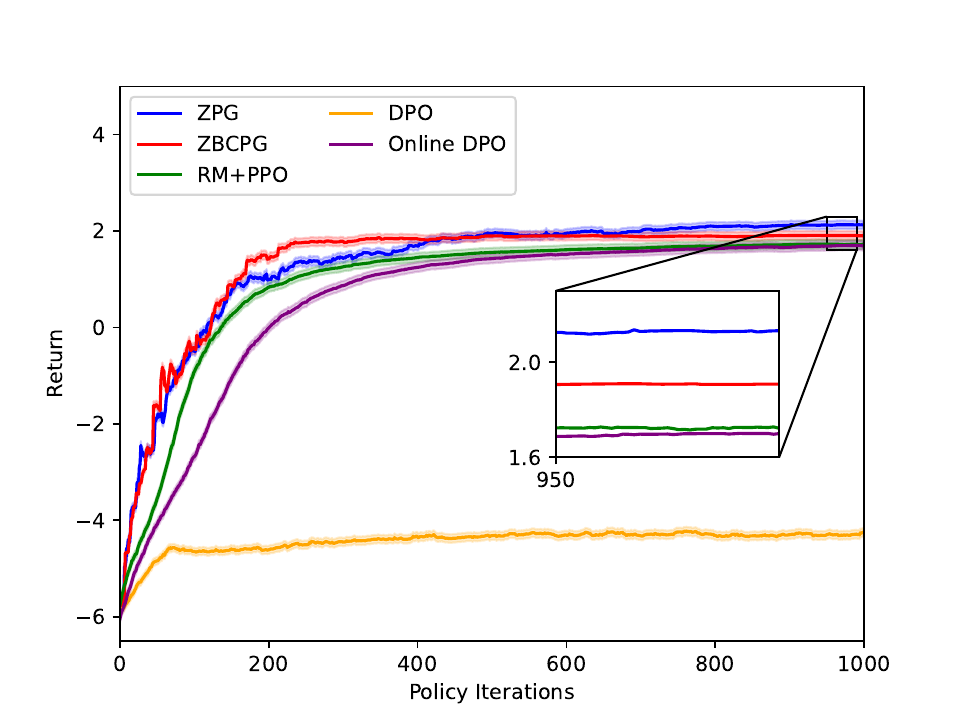}
         }
    \subfigure[Weibull, $M=200$]{
        \centering
         \includegraphics[width=0.48\linewidth]{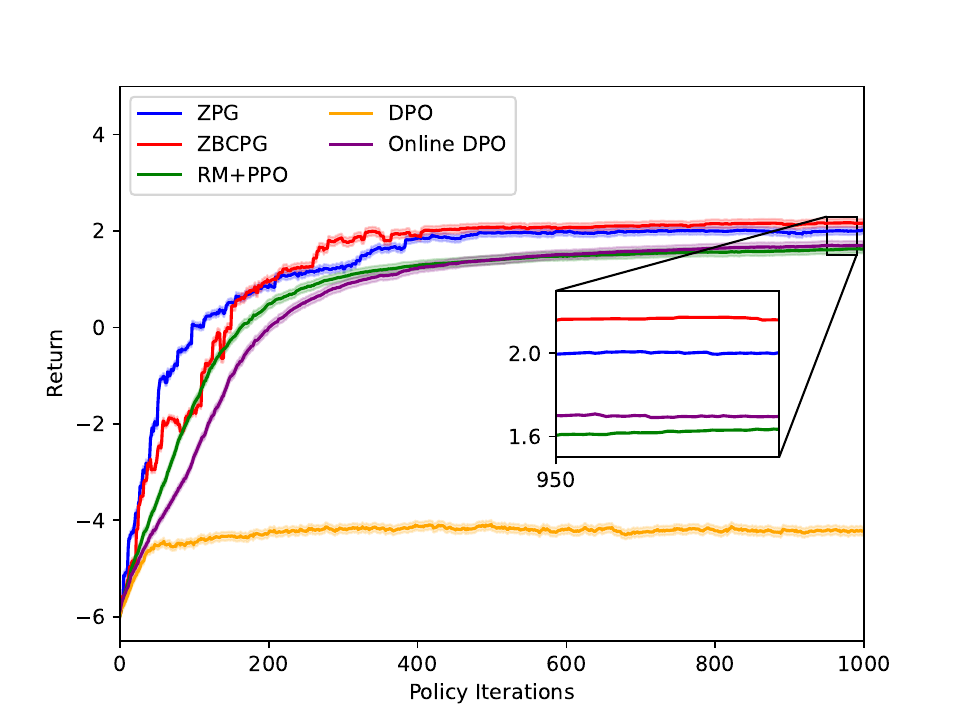}
         }
    \caption{GridWorld with Weibull Feedback: (a) the return of ZPG, ZBCPG, and RLHF baselines under Weibull human feedback with panel size $M=1000$, and (b) the trajectory return of ZPG, ZBCPG, and RLHF baselines under Weibull human feedback with panel size $M=200$. All results are averaged over $10^5$ repetitions of policy evaluation and shaded areas indicate confidence intervals.}
    \label{fig:weibull}
\end{figure}

\textbf{Influence to Return Comparison.}
The implementations of all algorithms are the same as in Sec.~\ref{sec:experiment} except that we use the Weibull link function in both ZPG and ZBCPG. The results are shown in Fig.~\ref{fig:weibull}(a) and Tab.~\ref{tab:weibull}. It can be seen that both our algorithms perform much better than the baselines in terms of convergence rate and the quality of the last iterate policy. Even though the trend is similar to the Bradley-Terry feedback results shown in Sec.~\ref{sec:experiment}, the performance gap is slightly larger due to the additional model mismatch factor. Both DPO and Online DPO are designed explicitly under the Bradley-Terry model, which is inaccurate for the ground-truth Weibull feedback, and the same applies to the reward model training step in the PPO baseline. However, the performance is not significantly degrading, which implies the Weibull link function is not fundamentally different than the Bradley-Terry model and allows these algorithms to have a compelling performance still. It is also shown that our algorithms are more flexible with the different human preference models and deliver better quality.
\begin{table}[ht]
    \centering
    \begin{tabular}{cccccc}
    \toprule
    Algorithm     & ZPG ({\color{blue} Ours}) & ZBCPG ({\color{blue} Ours}) & RM+PPO & DPO & Online DPO \\
    \midrule
    Return    &$\boldsymbol{2.13} \pm 0.09$ & $\boldsymbol{1.90} \pm 0.09$ & $1.72\pm 0.09$ & $-4.25 \pm 0.09$ & $1.69 \pm 0.09$ \\
    \bottomrule
    \end{tabular}
    \caption{Last Iterate Policy Average Return under Weibull Feedback with $M=1000$.}
    \label{tab:weibull}
\end{table}

\textbf{Panel Size.}
To study the influence of panel size $M$ on the quality of the learned policy, and to compare with our theoretical results presented in Theorem.~\ref{thm:ZPG} and Theorem~\ref{thm:ZBCPG}, we conduct the same experiment with Weibull preference feedback with the same setup as the previous subsection where we only use $M=100$ human evaluators for each trajectory pair, which is smaller than the $\gO(H^2)$ guarantee required by theory. The results are shown in Fig.~\ref{fig:weibull}(b) and Tab.~\ref{tab:weibull-small}. It can be seen that even with a much smaller human panel, our proposed algorithms can still converge to a local optimal policy which has a better quality than all benchmarks. Comparing the quality of the last iterate policy in Tab.~\ref{tab:weibull-small} to Tab.~\ref{tab:weibull}, it can also be deduced that the final policy found by our proposed algorithm is robust to the size of the human panel $M$, and the performance with less human evaluators does not significantly degenerate. However, it can also be seen from Fig.~\ref{fig:weibull} that both ZPG and ZBCPG have relatively larger policy quality fluctuation over the training trajectory compared to the three baselines, and this fluctuation increases as the number of human experts $M$ decreases, which poses stability concerns. This is partially because our proposed algorithms do not use KL regularization over the policies to constrain each gradient ascent step, which results in large policy alterations between iterations. We conjure that this is likely to be mitigated if we apply the KL regularization to our algorithms.
\begin{table}[ht]
    \centering
    \begin{tabular}{cccccc}
    \toprule
    Algorithm     & ZPG ({\color{blue} Ours}) & ZBCPG ({\color{blue} Ours}) & RM+PPO & DPO & Online DPO \\
    \midrule
    Return    &$\boldsymbol{2.02} \pm 0.08$ & $\boldsymbol{2.16} \pm 0.09$ & $1.63\pm 0.09$ & $-4.22 \pm 0.10$ & $1.69 \pm 0.09$ \\
    \bottomrule
    \end{tabular}
    \caption{Last Iterate Policy Average Return under Weibull Feedback with $M=100$.}
    \label{tab:weibull-small}
\end{table}

\section{Conclusion}
In this paper, we proposed two  RLHF algorithms without reward inference based on a zeroth-order policy gradient called ZPG and ZBCPG.
Both algorithms are shown to have a provable polynomial sample complexity to learn a stationary policy under mild conditions and exhibit nice empirical performances in environments with stochastic transitions, outperforming popular RLHF baselines.

\textbf{Limitations and Future Directions.} Both proposed algorithms require many online human queries for accurate gradient estimation to ensure fast convergence to a good policy, both in theory and in experiments, which may limit their practicality. Such limitation is common in all online RLHF algorithms, including online DPO~\citep{guo2024onlineDPO}. On the other hand, online RLHF algorithms exhibit better performances, which is also observed in our experiments, which may converge faster and require fewer iterations. One possible approach to avoid this limitation is to replace human feedback with AI feedback similar to~\citet{guo2024onlineDPO}. Another possible method is to replace the SGD style update in the current algorithm with a momentum-based optimization scheme, to reuse the human queries from previous policy iterations. The necessity of the current level of human queries will be further explored. Another limitation of our algorithms is the lack of strategic exploration, which may either fail to explore hard-to-reach trajectories or produce similar trajectories that are hard to provide preferences. One possible way to incorporate better exploration practice into our proposed algorithms is to replace the SGD type of policy update with projected SGD, which forces the agent to explore before convergence. Solving both limitations will be our future direction.

\section*{Acknowledgments}
The work of Qining Zhang and Lei Ying is supported in part by NSF under grants 2112471, 2134081, 2207548, 2240981, and 2331780.

\bibliography{ref,inlab-refs}
\bibliographystyle{unsrtnat}

\clearpage
\appendix

\section{Details of the Numerical Experiment in Sec. \ref{sec:experiment}}\label{sec:experiment-appendix}
In this section, we describe the implementation of our proposed algorithms and the implementation of baseline algorithms such as DPO, online DPO, and PPO.

\subsection{Experiments Comparing ZPG and ZPBCPG to Baselines}
In this section, we describe the implementation of algorithms in Fig.~\ref{fig:reward}(a), Fig.~\ref{fig:weibull}(a) and (b). For all algorithms, we tune the learning rates and perturbation distances if necessary. We also ensure the total number of trajectory samples and the total number of human queries are the same for fair comparison.

\textbf{ZPG.} We implemented ZPG according to Alg.~\ref{alg:ZPG} with trim size $\Delta = 0.001$ and batch size $N=1000$ for $T=1000$ iterations. After collecting $N$ trajectory pairs for the perturbed policy and the current policy, all panelists were queried to obtain the preference of all trajectory pairs.

\textbf{ZBCPG.} We implemented ZBCPG according to Alg.~\ref{alg:ZBCPG} with the same setup as ZPG with trim size $\Delta = 0.001$ and batch size $N=1000$ for $T=1000$ iterations. The coordinate batch size $K$ is chosen to be $5\times 4=20$, i.e., at each time we perturb the parameter $\xi_{s,a}$ associated with all actions for $5$ states simultaneously. After collecting $N = 1000$ trajectory pairs for the perturbed policy and the current policy, all panelists were queried to obtain the preference of all trajectory pairs. 

\textbf{RM+PPO.} The PPO baseline contains two components: reward model training and policy training. The reward model we consider used tabular parameterization. To train the reward model, we first used the randomly initialized policy to collect $5\times 10^5$ trajectory pairs, which is half the number of trajectories used in ZPG and ZBCPG. Then, we queried all panelists for the preferences of all trajectory pairs and obtained the population preference probability for all pairs. After that, we used SGD optimizer to train the reward model for $5$ epochs under the loss in~\citet{Christiano17RLHFTRPO}, which also used multiple human queries for each pair of trajectories. For the policy training, we used standard online PPO with a KL regularization weight being $0.1$. We also conducted $T = 1000$ policy iterations, and the agent collects $N = 1000$ trajectories between policy updates for improved training stability. The PPO loss in each iteration is optimized using SGD for $5$ epochs.

\textbf{DPO.} To train the DPO agent, we also conducted $T = 1000$ policy iterations, and the agent collects $N = 1000$ trajectory pairs between policy updates. The KL regularization weight in the DPO loss is $0.1$. For each pair of trajectories, the agent queried all panelists for preferences and then used the averaged preference probability to replace the single human preference in the vanilla DPO loss, similar to the loss in~\citet{Christiano17RLHFTRPO}, to ensure fair comparison among algorithms. The DPO loss at each iteration is optimized using SGD for $5$ epochs.

\textbf{Online DPO.} The implementation of online DPO is almost the same as offline DPO, except that we replaced the reference policy in the KL regularization with the current policy.

\subsection{Experiment Comparing ZPG to Distributed ZBCPG Implementation}
In this section, we describe the distributed implementation of ZBCPG, which produced the results in Fig.~\ref{fig:reward}(b). For distributed ZBCPG, we assume multiple parallel agents can collect trajectories and query humans in parallel, and the number of blocks in the legend of Fig.~\ref{fig:reward}(b) controls the level of parallelization. For example, when the number of blocks is $5$, we divide the parameters in the policy network into $5$ blocks. Then, five perturbed policies are created by perturbing each block individually. These policies are distributed to $5$ agents, and each agent will collect $N=1000$ trajectories simultaneously. However, to ensure fairness in the human panel query, we also divide the complete $M=1000$ people panel into $5$ sub-panels that could work in parallel, where each sub-panel includes $200$ panelists and will assist one agent with human preference queries. Therefore, the parallelized version of ZBCPG only has the advantage of sampling multiple trajectories, but not the access to more human experts. When the number of blocks is $5$, we perturbed the parameters associated with all actions for $5$ states simultaneously so each block contains $5\times 4 = 20$ parameters. Then, the $5$ blocks are distributed to $5$ agents to estimate the policy gradient. When the number of blocks is $25$, we perturbed parameters associated with all actions for a single state and each block contains $4$ parameters. When the number of blocks is $1$, we perturbed all parameters with Rademacher noise. The implementation of ZPG and all other parameters are the same as in the previous section.

\section{Proof Backbone for Theorems: Preference Estimation Error}
In the following sections, we provide proofs for both our main theorems, i.e., Theorem.~\ref{alg:ZPG} and Theorem.~\ref{alg:ZBCPG} for ZPG (Algorithm.~\ref{alg:ZPG}) and ZBCPG (Algorithm.~\ref{alg:ZBCPG}) respectively. Both algorithms follow the Lyapunov drift analysis framework where the key to the final result relies on a tight bound on the drift. At each iteration $t$ and for each trajectory pair, both algorithms estimate the population-level human preference probability $p_{t,n}$ and use it with the preference model in~\eqref{eq:preference-model} to estimate the reward difference between trajectories, which is later averaged to estimate the value function difference as in~\eqref{eq:preference-estimate}. Before presenting the proofs of the theorems, we first characterize the error incurred by preference estimation in the following Lemma:
\begin{lemma}[Concentration of Reward Difference]\label{lemma:conc-reward}
    Suppose $\Delta = \min\{\sigma(-H), 1-\sigma(H)\}$, for any trajectory pairs $(\tau_{n,0}, \tau_{n,1})$ that is queried from $M$ human evaluators. we have:
    \begin{align}
        &\E\left|\sigma^{-1}(p_{t,n}) - [r(\tau_{n,1}) - r(\tau_{n,0})]\right| \leq L \sqrt{\frac{2\log M}{M}} + \frac{2H}{M^2};\label{eq:conc-reward1} \\
        &\E\left|\sigma^{-1}(p_{t,n}) - [r(\tau_{n,1}) - r(\tau_{n,0})]\right|^2 \leq \frac{2 L^2 \log M}{M} + \frac{4H^2}{M^2}.\label{eq:conc-reward2}
    \end{align}
\end{lemma}
This lemma shows that our reward difference estimation from human preference is accurate as long as the number of human evaluators $M$ is large. The proof of this lemma is provided in Sec.~\ref{sec:proof-human}

\subsection{Proof of Lemma.~\ref{lemma:conc-reward}}\label{sec:proof-human}
Consider any iteration $t$ and any trajectory pairs $(\tau_{n,0}, \tau_{n,1})$ generated by the original and perturbed policy in both ZPG and ZBCPG. We first bound the estimation error of population-level human preference $\sP(\tau_{n,1} \succ \tau_{n,0})$ given any arbitrary trajectory pair in the following Lemma.~\ref{lemma:conc-win}. To formalize, let $\bar{o}_n$ be the empirical average over the human feedback $o_{n,m}$, i.e., 
$$
    \bar{o}_n = \sum_{m=1}^M \frac{o_{n,m}}{M}.
$$ 
Naturally, $\E[\bar{o}_n] = \E[o_{n,m}] = \sP(\tau_{n,1} \succ \tau_{n,0})$. And we need to characterize this estimation error first before we prove Lemma.~\ref{lemma:conc-reward}.
\begin{lemma}[Concentration of Preference Probability]\label{lemma:conc-win}
    For any $\delta<\frac{1}{4}$ and suppose $\Delta = \min\{\sigma(-H), 1-\sigma(H)\}$, for any trajectory pairs $(\tau_{n,0}, \tau_{n,1})$ that is queried from $M$ human evaluators, with probability at least $1-\delta$, we have:
    \begin{align*}
        \left|p_{t,n} - \sP(\tau_{n,1} \succ \tau_{n,0})\right| = \left|\trim\left[ \left.\bar{o}_n\right|\Delta\right] - \sP(\tau_{n,1}\succ \tau_{n,0})\right| \leq \sqrt{\frac{\log \frac{1}{\delta}}{M}},
    \end{align*}
    where $o_{n,m}$ is the feedback for the $m$-th human evaluator for this trajectory pair. 
\end{lemma}
\textbf{Proof.} 
First, we notice that for two trajectories $\tau_{n,1}$ and $\tau_{n,0}$, the reward difference of the two trajectories is bounded, i.e., $|r(\tau_{n,1}) - r(\tau_{n,0})| \leq H$, which implies the population-level preference probability $\sP(\tau_{n,1} \succ \tau_{n,0})$ is also bounded, i.e., 
$$
    \sP(\tau_{n,1} \succ \tau_{n,0}) \in [\sigma(-H), \sigma(H)] \subset [\Delta, 1-\Delta].
$$ 
Note that $\bar{o}_n$ may not be within $[\Delta, 1-\Delta]$ due to the finite number of human evaluators. Overall, we may have the following three cases:
\begin{enumerate}
    \item $\bar{o}_n \in [\Delta, 1-\Delta]$: we have $|\trim[\bar{o}_n|\Delta] - \sP(\tau_{n,1} \succ \tau_{n,0})| = |\bar{o}_n - \E[\bar{o}_n]|$.
    \item $\bar{o}_n > 1-\Delta \geq \sP(\tau_{n,1} \succ \tau_{n,0})$: we have $$|\trim[\bar{o}_n|\Delta]- \sP(\tau_{n,1} \succ \tau_{n,0})| = 1-\Delta- \sP(\tau_{n,1} \succ \tau_{n,0}) \leq |\bar{o}_n - \E[\bar{o}_n]|.$$
    \item $\bar{o}_n < \Delta \leq \sP(\tau_{n,1} \succ \tau_{n,0})$: we have $$|\trim[\bar{o}_n|\Delta]- \sP(\tau_{n,1} \succ \tau_{n,0})| = \sP(\tau_{n,1} \succ \tau_{n,0}) - \Delta \leq |\bar{o}_n - \E[\bar{o}_n]|.$$
\end{enumerate}
To summarize, we have $|\trim[\bar{o}_n|\Delta]- \sP(\tau_{n,1}\succ \tau_{n,0})| \leq |\bar{o}_n - \E[\bar{o}_n]|$. According to Hoeffding's inequality for Bernoulli random variable, we have:
\begin{align*}
    \sP\left( |\trim[\bar{o}_n|\Delta]- \sP(\tau_{n,1} \succ \tau_{n,0})| > \sqrt{\frac{\log \frac{1}{\delta}}{M}} \right) 
    \leq &  \sP\left( |\bar{o}_n - \E[\bar{o}_n]| > \sqrt{\frac{\log \frac{1}{\delta}}{M}} \right) \\
    \leq & \exp\left( -2M \frac{\log \frac{1}{\delta}}{M}\right)
    \leq \delta,
\end{align*}
which concludes the proof.$\hfill\blacksquare$

Next, we are ready to bound the error between the reward difference estimator after plugging in the preference probability estimator $p_{t,n}$ to the inverse link function $\sigma^{-1}$ in Lemma.~\ref{lemma:conc-reward}. We first prove~\eqref{eq:conc-reward1} and then prove~\eqref{eq:conc-reward2}. Let $\gE_n$ be the concentration event that the event in Lemma.~\ref{lemma:conc-win} holds. By Lemma.~\ref{lemma:conc-win}, when the concentration event does not hold, we can bound the term in the expectation of the left-hand side as:
\begin{align*}
    \left|\sigma^{-1}(p_{t,n}) - [r(\tau_{n,1}) - r(\tau_{n,0})]\right| \leq 2 H.
\end{align*}
When the concentration event holds, we first notice that $r(\tau_1) - r(\tau_0) = \sigma^{-1}(\sP(\tau_{n,1} \succ \tau_{n,0}))$. By definition, both $p_{t,n}$ and $\sP(\tau_{n,1} \succ \tau_{n,0})$ belongs to the interval $[\Delta, 1-\Delta]$. This allows us to use assumption~\ref{assumpt:linkgrad} and we have:
\begin{align*}
    \left|\sigma^{-1}(p_{t,n}) - [r(\tau_{n,1}) - r(\tau_{n,0})]\right| 
    = & \left|\sigma^{-1}(p_{t,n}) - \sigma^{-1}(\sP(\tau_{n,1} \succ \tau_{n,0}))\right|\\
    \leq & L \left| p_{t,n} - \sP(\tau_{n,1} \succ \tau_{n,0})) \right|\\
    \leq & L \sqrt{\frac{\log \frac{1}{\delta}}{M}},
\end{align*}
where the last inequality uses the smoothness of the inverse link function, i.e., Lemma.~\ref{lemma:conc-win}. Combining both cases and let $\delta = 1/M^2$, we have:
\begin{align*}
    \E\left|\sigma^{-1}(p_{t,n}) - [r(\tau_{n,1}) - r(\tau_{n,0})]\right| \leq & \sP(\gE_n^\complement) L \sqrt{\frac{\log \frac{1}{\delta}}{M}} + \sP(\gE_n) 2 H
    \leq  L \sqrt{\frac{2\log M}{M}} + \frac{2H}{M^2},
\end{align*}
which concludes the proof of the first inequality. For the second inequality, we apply a similar procedure. Suppose the concentration event $\gE_n$ does not hold, then we can bound the term as:
\begin{align*}
    \left|\sigma^{-1}(p_{t,n}) - [r(\tau_{n,1}) - r(\tau_{n,0})]\right|^2 \leq 4H^2.
\end{align*}
This is because the inverse link function is bounded after applying the trim operator to the preference probability estimation.
When the concentration holds, we also employ the continuity of the inverse link function as:
\begin{align*}
    \left|\sigma^{-1}(p_{t,n}) - [r(\tau_{n,1}) - r(\tau_{n,0})]\right|^2 
    \leq  \left(L \sqrt{\frac{\log \frac{1}{\delta}}{M}}\right)^2 = \frac{L^2 \log \frac{1}{\delta}}{M},
\end{align*}
Combining both cases and let $\delta = 1/M^2$, we have:
\begin{align*}
    \E\left|\sigma^{-1}(p_{t,n}) - [r(\tau_{n,1}) - r(\tau_{n,0})]\right|^2 \leq & \sP(\gE_n^\complement) \frac{2L^2 \log M }{M} + \sP(\gE_n) 4H^2
    \leq  \frac{2 L^2 \log M}{M} + \frac{4H^2}{M^2},
\end{align*}
which concludes the proof of both inequalities. 

\section{Proof of Theorem.~\ref{thm:ZPG}}
Recall that we choose the smoothed value function $V_\mu(\pi_{\vtheta})$ as the Lyapunov function, i.e.,
\begin{align*}
    V_\mu(\pi_{\vtheta}) = \E_{\vv'}\left[ V(\pi_{\vtheta + \mu \vv'}) \right],
\end{align*}
where the vector $\vv'$ is sampled from a $d$-dimensional uniform distribution over a unit ball. We first provide properties of the smoothed function, which is proved in~\citet[Lemma. 1]{liu18ZOSVRG}.
\begin{lemma}\label{lemma:smoothfunction}
    Suppose $\vv$ is sampled from a uniform distribution over a unit sphere and $\vv'$ is sampled from a uniform distribution over a unit ball both in $d$-dimensional space and Assumption.~\ref{assumpt:smooth} holds. Then the smoothed value function $V_\mu(\pi_{\vtheta})$ defined above satisfies:
    \begin{enumerate}
        \item $V_\mu(\pi_{\vtheta})$ is $L$-smooth and satisfies: 
        \begin{align}\label{eq:smooth-grad-def}
            \nabla_{\vtheta} V_\mu(\pi_{\vtheta}) = \E_{\vv}\left[ \frac{d}{\mu} \left( V(\pi_{\vtheta + \mu \vv}) - V(\pi_{\vtheta}) \right) \vv \right].
        \end{align}
        \item For any $\vtheta \in \sR^d$, the function value difference satisfies:
        \begin{align}\label{eq:smooth-value}
            \left| V_\mu(\pi_{\vtheta}) - V(\pi_{\vtheta}) \right| \leq \frac{L\mu^2}{2}.
        \end{align}
        \item For any $\vtheta \in \sR^d$, the gradient difference satisfies:
        \begin{align}\label{eq:smooth-grad}
            &\left\| \nabla_{\vtheta} V_\mu(\pi_{\vtheta}) - \nabla_{\vtheta} V(\pi_{\vtheta}) \right\|_2 \leq \frac{\mu L d}{2}.
        \end{align}
        \item For any $\vtheta \in \sR^d$, the gradient noise satisfies:
        \begin{align}\label{eq:smooth-grad-noise}
            &\E_{\vv}\left[\left\| \frac{d}{\mu} \left( V(\pi_{\vtheta + \mu \vv}) - V(\pi_{\vtheta}) \right) \vv \right\|_2^2\right] \leq 2d \left\| \nabla_{\vtheta} V(\pi_{\vtheta}) \right\|_2^2 + \frac{\mu^2 L^2 d^2}{2}.
        \end{align}
    \end{enumerate}
\end{lemma}
Now, we are ready to prove Theorem.~\ref{thm:ZPG}. We choose $V_\mu(\pi_{\vtheta})$ to be the Lyapunov function, where $\boldsymbol{v}$ is sampled from a unit ball. We first analyze its drift, from the smoothness of $V_\mu(\pi_{\vtheta})$ by Lemma.~\ref{lemma:smoothfunction}, we have the following upper bound:
\begin{align*}
    V_\mu(\pi_{\vtheta_t}) - V_\mu(\pi_{\vtheta_{t+1}}) \leq& \langle - \nabla_{\vtheta} V_\mu(\pi_{\vtheta_t}), \vtheta_{t+1} - \vtheta_t\rangle + \frac{L}{2} \|\vtheta_{t+1} - \vtheta_t\|_2^2 \\
    =& -\alpha\langle \nabla_{\vtheta} V_\mu(\pi_{\vtheta_t}), \hat{\vg}_t\rangle + \frac{L\alpha^2}{2} \|\hat{\vg}_t\|_2^2\\
    =& -\alpha\| \nabla_{\vtheta} V_\mu(\pi_{\vtheta_t}) \|_2^2 + \alpha \langle \nabla_{\vtheta} V_\mu(\pi_{\vtheta_t}), \nabla_{\vtheta} V_\mu(\pi_{\vtheta_t}) - \hat{\vg}_t\rangle + \frac{L\alpha^2}{2} \|\hat{\vg}_t\|_2^2.
\end{align*}
We bound the three terms above one by one, the first term can be bounded using~\eqref{eq:smooth-grad-def} from Lemma.~\ref{lemma:smoothfunction}. We first have:
\begin{align*}
    \| \nabla_{\vtheta} V(\pi_{\vtheta_t}) \|_2^2 
    = & \| \nabla_{\vtheta} V(\pi_{\vtheta_t}) - \nabla_{\vtheta} V_\mu(\pi_{\vtheta_t}) + \nabla_{\vtheta} V_\mu(\pi_{\vtheta_t}) \|_2^2\\
    \leq & 2\| \nabla_{\vtheta} V_\mu(\pi_{\vtheta_t}) - \nabla_{\vtheta} V(\pi_{\vtheta_t}) \|_2^2 + 2\| \nabla_{\vtheta} V_\mu(\pi_{\vtheta_t}) \|_2^2\\
    \leq & 2\| \nabla_{\vtheta} V_\mu(\pi_{\vtheta_t}) \|_2^2 + \frac{\mu^2 L^2 d^2}{2},
\end{align*}
where the first inequality uses $(a+b)^2 \leq 2a^2 + 2b^2$ and the second inequality uses~\eqref{eq:smooth-grad}. This implies:
\begin{align*}
    \| \nabla_{\vtheta} V_\mu(\pi_{\vtheta_t}) \|_2^2 \geq \frac{1}{2}\| \nabla_{\vtheta} V(\pi_{\vtheta_t}) \|_2^2 - \frac{\mu^2 L^2 d^2}{4}.
\end{align*}
And we have a bound on the negative drift term as:
\begin{align*}
    -\alpha\| \nabla_{\vtheta} V_\mu(\pi_{\vtheta_t}) \|_2^2 \leq -\frac{\alpha}{2} \| \nabla_{\vtheta} V(\pi_{\vtheta_t}) \|_2^2 + \frac{\alpha\mu^2 L^2 d^2}{4}.
\end{align*}
We take a conditional expectation of the drift over the natural filtration $\gF_t$ of time $t$ and obtain:
\begin{equation}\label{eq:ZPGproof-drift}
\begin{aligned}
    &\E\left[ V_\mu(\pi_{\vtheta_t}) - V_\mu(\pi_{\vtheta_{t+1}}) | \gF_t\right]\\
    \leq & -\frac{\alpha}{2} \| \nabla_{\vtheta} V(\pi_{\vtheta_t}) \|_2^2   + \frac{\alpha\mu^2 L^2 d^2}{4} + \frac{L\alpha^2}{2} \underbrace{\E\left[\left.\|\hat{\vg}_t\|_2^2 \right| \gF_t \right]}_{\mathsf{Var}_t}
    + \alpha \underbrace{\langle \nabla_{\vtheta} V_\mu(\pi_{\vtheta_t}), \nabla_{\vtheta} V_\mu(\pi_{\vtheta_t}) - \E\left[\left.\hat{\vg}_t \right| \gF_t\right]\rangle}_{\mathsf{Bias}_t}.
\end{aligned}
\end{equation}
Then, we analyze the two positive drift terms $\mathsf{Bias}_t$ for the first-order gradient bias and $\mathsf{Var}_t$ for the gradient noise separately and then construct a tight upper bound of the Lyapunov drift. The results are summarized in the following two lemmas where the proofs are deferred to Sec.~\ref{sec:proofZPGbias} and Sec.~\ref{sec:proofZPGvar}:
\begin{lemma}\label{lemma:ZPGbias}
    For ZPG and let $M\geq 8 (H/L)^{\frac{2}{3}}$, conditioned on the information filtration $\gF_t$ of any time $t$, the gradient bias can be upper bounded as follows:
    \begin{align*}
        \mathsf{Bias}_t =& \langle \nabla_{\vtheta} V_\mu(\pi_{\vtheta_t}), \nabla_{\vtheta} V_\mu(\pi_{\vtheta_t}) - \E\left[\left.\hat{\vg}_t \right| \gF_t\right]\rangle
        \leq  \frac{2dL}{\mu}\sqrt{\frac{\log M}{M}}\left\|\nabla_{\vtheta} V(\pi_{\vtheta_t}) \right\|_2 + d^2L^2\sqrt{\frac{\log M}{M}}.
    \end{align*}
\end{lemma}
\begin{lemma}\label{lemma:ZPGvar}
    For ZPG and let $M\geq 4 (H/L)^{2}$, conditioned on the information filtration $\gF_t$ of any time $t$, the gradient bias can be upper bounded as follows:
    \begin{align*}
        \mathsf{Var}_t =  \E\left[\left.\|\hat{\vg}_t\|_2^2 \right| \gF_t \right]  \leq 6d \left\| \nabla_{\vtheta} V(\pi_{\vtheta_t}) \right\|_2^2 + \frac{3\mu^2 L^2 d^2}{2} + \frac{9 d^2 L^2 \log M}{\mu^2 M} + \frac{12 d^2 H^2}{\mu^2 N}.
    \end{align*}
\end{lemma}
Now we use Lemma.~\ref{lemma:ZPGbias} and Lemma.~\ref{lemma:ZPGvar} and combine the drift bound in~\eqref{eq:ZPGproof-drift} and obtain:
\begin{align*}
    &\E\left[ V_\mu(\pi_{\vtheta_t}) - V_\mu(\pi_{\vtheta_{t+1}}) | \gF_t\right]\\
    \leq & -\frac{\alpha}{2} \| \nabla_{\vtheta} V(\pi_{\vtheta_t}) \|_2^2  + \frac{\alpha\mu^2 L^2 d^2}{4} + \alpha \mathsf{Bias}_t + \frac{L\alpha^2}{2} \mathsf{Var}_t \\
    \leq & -\frac{\alpha}{2} \| \nabla_{\vtheta} V(\pi_{\vtheta_t}) \|_2^2  + \frac{\alpha\mu^2 L^2 d^2}{4}+ \frac{2\alpha dL}{\mu} \sqrt{\frac{\log M}{M}} \left\|\nabla_{\vtheta} V(\pi_{\vtheta_t}) \right\|_2 + \alpha d^2L^2\sqrt{\frac{\log M}{M}}\\
    &+ 3\alpha^2dL \left\| \nabla_{\vtheta} V(\pi_{\vtheta_t}) \right\|_2^2 + {\alpha^2\mu^2 L^3 d^2} + \frac{5 \alpha^2 d^2 L^3 \log M}{\mu^2 M} + \frac{6 \alpha^2 d^2 H^2 L}{\mu^2 N}.
\end{align*}
Let $\alpha \leq (12dL)^{-1}$, we can further simplify the drift bound as:
\begin{align*}
    &\E\left[ V_\mu(\pi_{\vtheta_t}) - V_\mu(\pi_{\vtheta_{t+1}}) | \gF_t\right] \\
    \leq& -\frac{\alpha}{2} \| \nabla_{\vtheta} V(\pi_{\vtheta_t}) \|_2^2  + \frac{\alpha\mu^2 L^2 d^2}{4}+ \frac{2\alpha dL}{\mu} \sqrt{\frac{\log M}{M}} \left\|\nabla_{\vtheta} V_\mu(\pi_{\vtheta_t}) \right\|_2\\
    & + \alpha d^2L^2\sqrt{\frac{\log M}{M}}+ \frac{1}{4}\alpha \left\| \nabla_{\vtheta} V(\pi_{\vtheta_t}) \right\|_2^2 + {\alpha \mu^2 L^2 d} + \frac{5 \alpha^2 d^2 L^3 \log M}{\mu^2 M} + \frac{6 \alpha^2 d^2 H^2 L}{\mu^2 N}\\
    \leq & -\frac{\alpha}{4} \| \nabla_{\vtheta} V(\pi_{\vtheta_t}) \|_2^2  + \frac{\alpha\mu^2 L^2 d^2}{2}+ \frac{2\alpha dL}{\mu} \sqrt{\frac{\log M}{M}} \left\|\nabla_{\vtheta} V_\mu(\pi_{\vtheta_t}) \right\|_2 + \alpha d^2L^2\sqrt{\frac{\log M}{M}}\\
    &+ \frac{5 \alpha^2 d^2 L^3 \log M}{\mu^2 M} + \frac{6 \alpha^2 d^2 H^2 L}{\mu^2 N},
\end{align*}
where the last inequality merges both positive and negative drifts on the square of the gradient together and assumes $d\geq 4$. When $\left\|\nabla_{\vtheta} V_\mu(\pi_{\vtheta_t}) \right\|_2 $ is large, typically, when we have:
\begin{align*}
    \left\|\nabla_{\vtheta} V_\mu(\pi_{\vtheta_t}) \right\|_2 \geq \frac{16 d L}{\mu} \sqrt{\frac{\log M}{M}},
\end{align*}
we have the positive first-order drift regarding the gradient norm is bounded by:
\begin{align*}
    \frac{2\alpha dL}{\mu} \sqrt{\frac{\log M}{M}} \left\|\nabla_{\vtheta} V_\mu(\pi_{\vtheta_t}) \right\|_2 \leq \frac{\alpha}{8}\left\|\nabla_{\vtheta} V_\mu(\pi_{\vtheta_t}) \right\|_2^2, 
\end{align*}
and thus can be merged into the negative drift as follows:
\begin{align*}
    &\E\left[ V_\mu(\pi_{\vtheta_t}) - V_\mu(\pi_{\vtheta_{t+1}}) | \gF_t\right] \\
    \leq & -\frac{\alpha}{8} \| \nabla_{\vtheta} V(\pi_{\vtheta_t}) \|_2^2  + \frac{\alpha\mu^2 L^2 d^2}{2}+ \alpha d^2L^2\sqrt{\frac{\log M}{M}}
    + \frac{5 \alpha^2 d^2 L^3 \log M}{\mu^2 M} + \frac{6 \alpha^2 d^2 H^2 L}{\mu^2 N}.
\end{align*}
If on the other hand, the gradient is small, i.e., we have:
\begin{align*}
    \left\|\nabla_{\vtheta} V_\mu(\pi_{\vtheta_t}) \right\|_2 \leq \frac{16 d L}{\mu} \sqrt{\frac{\log M}{M}},
\end{align*}
and then we can upper bound the first-order drift regarding gradient norm in another way:
\begin{align*}
    \frac{2\alpha dL}{\mu} \sqrt{\frac{\log M}{M}} \left\|\nabla_{\vtheta} V_\mu(\pi_{\vtheta_t}) \right\|_2 \leq \frac{32\alpha d^2L^2 \log M}{\mu^2 M}.
\end{align*}
So in this case the total drift can be upper bounded as:
\begin{align*}
    &\E\left[ V_\mu(\pi_{\vtheta_t}) - V_\mu(\pi_{\vtheta_{t+1}}) | \gF_t\right] \\
    \leq & -\frac{\alpha}{4} \| \nabla_{\vtheta} V(\pi_{\vtheta_t}) \|_2^2  +  \frac{\alpha\mu^2 L^2 d^2}{2} + \frac{32\alpha d^2L^2 \log M}{\mu^2 M} + \alpha d^2L^2\sqrt{\frac{\log M}{M}}
    + \frac{5 \alpha^2 d^2 L^3 \log M}{\mu^2 M} + \frac{6 \alpha^2 d^2 H^2 L}{\mu^2 N}\\
    \leq & -\frac{\alpha}{4} \| \nabla_{\vtheta} V(\pi_{\vtheta_t}) \|_2^2   + \frac{\alpha\mu^2 L^2 d^2}{2} + \frac{33\alpha d^2L^2 \log M}{\mu^2 M} + \alpha d^2L^2\sqrt{\frac{\log M}{M}}+ \frac{6 \alpha^2 d^2 H^2 L}{\mu^2 N}.
\end{align*}
The last inequality uses $\alpha \leq (12dL)^{-1}$ and $d\geq4$. So by taking a maximum of both bounds and taking an expectation, we have:
\begin{align*}
    &\E\left[ V_\mu(\pi_{\vtheta_t})\right] - \E\left[ V_\mu(\pi_{\vtheta_{t+1}})\right] \\
    \leq & -\frac{\alpha}{8} \E\left[ \| \nabla_{\vtheta} V(\pi_{\vtheta_t}) \|_2^2\right]  + \frac{33\alpha d^2L^2 \log M}{\mu^2 M} + \frac{\alpha\mu^2 L^2 d^2}{2}+ \alpha d^2L^2\sqrt{\frac{\log M}{M}}+ \frac{6 \alpha^2 d^2 H^2 L}{\mu^2 N}.
\end{align*}
We then take a telescoping sum which results in:
\begin{align*}
    &\E\left[ V_\mu(\pi_{\vtheta_0})\right] - \E\left[ V_\mu(\pi_{\vtheta_T})\right]\\
    \leq &  -\frac{\alpha}{8} \sum_{t=0}^{T-1}\E\left[ \| \nabla_{\vtheta} V(\pi_{\vtheta_t}) \|_2^2\right]
    + \left(\frac{33\alpha d^2L^2 \log M}{\mu^2 M} + \frac{\alpha\mu^2 L^2 d^2}{2}+ \alpha d^2L^2\sqrt{\frac{\log M}{M}}+ \frac{6 \alpha^2 d^2 H^2 L}{\mu^2 N} \right)T.
\end{align*}
We choose $\mu$ to balance the terms inside the parenthesis. Specifically, let
\begin{align*}
    \mu^2 = \max\left\{ 9\sqrt{\frac{\log M}{M}}, \frac{4 H}{L\sqrt{dN}} \right\}, \quad \alpha = \frac{1}{12 d L}.
\end{align*}
Then, we have the following inequality on the positive drift:
\begin{align*}
    \frac{\alpha\mu^2 L^2 d^2}{2} \geq \frac{33\alpha d^2L^2 \log M}{\mu^2 M}, \quad \frac{\alpha\mu^2 L^2 d^2}{2} \geq \frac{6 \alpha^2 d^2 H^2 L}{\mu^2 N}.
\end{align*}
So we have the final results in Theorem.~\ref{thm:ZPG} as:
\begin{align*}
    \frac{1}{T}\sum_{t=0}^{T-1}\E\left[ \| \nabla_{\vtheta} V(\pi_{\vtheta_t}) \|_2^2\right] 
    \leq & \frac{8\left( \E\left[ V_\mu(\pi_{\vtheta_0})\right] - \E\left[ V_\mu(\pi_{\vtheta_T})\right] \right)}{T\alpha} + 12\mu^2 L^2 d^2 + 8d^2L^2\sqrt{\frac{\log M}{M}}\\
    \leq & \frac{8\left( \E\left[ V(\pi_{\vtheta_0})\right] - \E\left[ V(\pi_{\vtheta_T})\right] \right)}{T\alpha} + \frac{8\mu^2 L}{T\alpha} + 12\mu^2 L^2 d^2 + 8d^2L^2\sqrt{\frac{\log M}{M}}\\
    =& \gO\left( \frac{H Ld}{T} + \frac{d^2 L^2 \sqrt{\log M}}{\sqrt{M}} + \frac{HL d\sqrt{d}}{\sqrt{N}} \right).
\end{align*}

\subsection{Proof of Lemma.~\ref{lemma:ZPGbias}}\label{sec:proofZPGbias}

We analyze the first-order gradient bias term $\mathsf{Bias}_t$ in the drift of~\eqref{eq:ZPGproof-drift}. Notice that conditioned over $\gF_t$, the randomness only comes from sampling the perturbation direction, sampling trajectories, and obtaining human feedback. By~\eqref{eq:smooth-grad-def} and the definition of $\hat{\vg}_t$, we have:
\begin{align*}
    \E\left[\nabla_{\vtheta} V_\mu(\pi_{\vtheta_t}) - \left.\hat{\vg}_t \right| \gF_t\right]
    =& \E\left[ \left.\frac{d\left( V(\pi_{\vtheta_t + \mu \vv_t}) - V(\pi_{\vtheta_t}) \right)}{\mu}  \vv_t \right| \gF_t\right] - \E\left[\left. \frac{d}{ \mu }\frac{\sum_{n=1}^N \sigma^{-1}(p_{t,n})}{N} \vv_t \right| \gF_t \right].
\end{align*}
Since each trajectory pair $(\tau_{n,0}, \tau_{n,1})$ is generated independently from other trajectories, we have:
\begin{align*}
    &\E\left[\left. \frac{d}{ \mu }\frac{\sum_{n=1}^N \sigma^{-1}(p_{t,n})}{N} \vv_t \right| \gF_t \right]\\
    =&\frac{d}{ \mu } \E\left[\left. \E\left[\left. \frac{\sum_{n=1}^N \sigma^{-1}(p_{t,n})}{N}  \right| \vv_t \right] \vv_t \right| \gF_t \right]\\
    =&\frac{d}{ \mu } \E\left[\left. \E\left[\left. \sigma^{-1}(p_{t,n})  \right| \vv_t \right] \vv_t \right| \gF_t \right]\\
    =&\frac{d}{ \mu } \E\left[\left. \E\left[\left. \left( r(\tau_{n,1}) - r(\tau_{n,0}) \right) \right| \vv_t \right] \vv_t \right| \gF_t \right]- \frac{d}{ \mu }\E\left[\left. \E\left[\left. \left( \sigma^{-1}(p_{t,n}) - [r(\tau_{n,1}) - r(\tau_{n,0})] \right) \right| \vv_t \right] \vv_t \right| \gF_t \right],
\end{align*}
where the first equality uses the law of total expectation. The second equality uses the independent nature of the trajectories generated by the same policy pair. In the last inequality, we aim to bound the difference between the reward difference estimate from human preference and the ground-truth reward difference, which can be constructed through concentration inequalities. Substitute it back into the bias, we have:
\begin{align*}
    \E\left[\nabla_{\vtheta} V_\mu(\pi_{\vtheta_t}) - \left.\hat{\vg}_t \right| \gF_t\right]
    =& \frac{d}{\mu}\E\left[\left. \E\left[ \left.\left(  V(\pi_{\vtheta_t + \mu \vv_t}) - V(\pi_{\vtheta_t})  -  r(\tau_{n,1}) + r(\tau_{n,0}) \right) \right| \vv_t \right]  \vv_t \right| \gF_t\right]\\
    &+\frac{d}{ \mu } \E\left[\left. \E\left[\left. \left( \sigma^{-1}(p_{t,n}) - [r(\tau_{n,1}) - r(\tau_{n,0})] \right) \right| \vv_t \right] \vv_t \right| \gF_t \right],
\end{align*}
Notice that in the inner expectation of the first term, the randomness comes from sampling trajectories given both policies $\pi_{\vtheta_t + \mu\vv_t}$ and $\pi_{\vtheta_t}$, we have:
\begin{align*}
    &\E\left[\left(  V(\pi_{\vtheta_t + \mu \vv_t}) - V(\pi_{\vtheta_t})  -  r(\tau_{n,1}) + r(\tau_{n,0}) \right) | \vv_t\right]\\
    =& V(\pi_{\vtheta_t + \mu \vv_t}) - \E\left[r(\tau_{n,1})| \tau_{n,1} \sim  \pi_{\vtheta_t + \mu \vv_t}\right] + \E\left[r(\tau_{n,0})| \tau_{n,0} \sim  \pi_{\vtheta_t}\right] - V(\pi_{\vtheta_t})
    =0.
\end{align*}
The last equality is due to the definition of the value function where trajectory returns are unbiased estimates of the value functions and the fact that $\tau_{n,1}$ and $\tau_{n,0}$ are generated by the perturbed policy and the original policy independently. Then, the bias drift only comes from the preference estimation error and we will be able to plug it back and have:
\begin{align*}
    \mathsf{Bias}_t 
    = &\langle \nabla_{\vtheta} V_\mu(\pi_{\vtheta_t}), \nabla_{\vtheta} V_\mu(\pi_{\vtheta_t}) - \E\left[\left.\hat{\vg}_t \right| \gF_t\right]\rangle \\
    = & \frac{d}{\mu} \E\left[\left(\sigma^{-1}(p_{t,n}) -  [r(\tau_{n,1}) - r(\tau_{n,0})] \right) \langle \nabla_{\vtheta} V_\mu(\pi_{\vtheta_t}),\vv_t \rangle | \gF_t\right]\\
    \leq & \frac{d}{\mu} \left\|\nabla_{\vtheta} V_\mu(\pi_{\vtheta_t}) \right\|_2 \E\left[\left. \E\left[ \left.\left| \sigma^{-1}(p_{t,n}) -  [r(\tau_{n,1}) - r(\tau_{n,0})]  \right| \right| \tau_{n,1}, \tau_{n,0} \right] \right| \gF_t\right]\\
    \leq & \frac{d}{\mu} \left\|\nabla_{\vtheta} V_\mu(\pi_{\vtheta_t}) \right\|_2 \left(L\sqrt{\frac{2\log M}{M}} + \frac{2H}{M^2}\right)\\
    \leq& \frac{2dL}{\mu} \left\|\nabla_{\vtheta} V_\mu(\pi_{\vtheta_t}) \right\|_2 \sqrt{\frac{\log M}{M}}\\
    \leq & \frac{2dL}{\mu}\sqrt{\frac{\log M}{M}}\left\|\nabla_{\vtheta} V(\pi_{\vtheta_t}) \right\|_2 + d^2L^2\sqrt{\frac{\log M}{M}},
\end{align*}
where the first inequality uses Cauchy-Schwarz inequality and the second inequality uses Lemma.~\ref{lemma:conc-reward}. The second last inequality is obtained by choosing $M \geq 8 (H/L)^{\frac{2}{3}}$. The final inequality is obtained by applying the third property of Lemma.~\ref{lemma:smoothfunction}.

\subsection{Proof of Lemma.~\ref{lemma:ZPGvar}}\label{sec:proofZPGvar}
We aim for the variance term $\mathsf{Var}_t$ in the Lyapunov drift in~\eqref{eq:ZPGproof-drift}. We first have upper bound variance as follows:
\begin{align*}
    \mathsf{Var}_t 
    = & \E\left[\left.\left\|\frac{d}{ \mu }\frac{\sum_{n=1}^N \sigma^{-1}(p_{t,n})}{N} \vv_t \right\|_2^2 \right| \gF_t \right]\\
    \leq & 3 \underbrace{\E\left[\left.\left\|\frac{d \sum_{n=1}^N \left(\sigma^{-1}(p_{t,n}) - [r(\tau_{n,1}) - r(\tau_{n,0})] \right) }{ \mu N}\vv_t \right\|_2^2 \right| \gF_t \right]}_{\mathsf{Var}_{t,1}}\\
    &+ 3 \underbrace{\E\left[\left.\left\| \frac{d}{\mu} \left(\frac{\sum_{n=1}^N \left(r(\tau_{n,1}) - r(\tau_{n,0}) \right) }{N} - \left( V(\pi_{\vtheta_t + \mu \vv_t}) - V(\pi_{\vtheta_t}) \right)\right)\vv_t \right\|_2^2 \right| \gF_t \right]}_{\mathsf{Var}_{t,2}}\\
    & + 3\underbrace{\E\left[\left.\left\| \frac{d}{\mu}  \left( V(\pi_{\vtheta_t + \mu \vv_t}) - V(\pi_{\vtheta_t}) \right)\vv_t \right\|_2^2 \right| \gF_t \right]}_{\mathsf{Var}_{t,3}}.
\end{align*}
The first term comes from the human preference estimation error which can be bounded using \eqref{eq:conc-reward2} from Lemma.~\ref{lemma:conc-reward} as follows:
\begin{align*}
    \mathsf{Var}_{t,1} = &\E\left[\left.\left\|\frac{d \sum_{n=1}^N \left(\sigma^{-1}(p_{t,n}) - [r(\tau_{n,1}) - r(\tau_{n,0})] \right) }{ \mu N}\vv_t \right\|_2^2 \right| \gF_t \right]\\
    = & \frac{d^2}{\mu^2 N^2} \E \left[\left|\sum_{n=1}^N \left(\sigma^{-1}(p_{t,n}) - [r(\tau_{n,1}) - r(\tau_{n,0})] \right) \right|^2 \right]\\
    \leq & \frac{d^2}{\mu^2 N} \sum_{n=1}^N \E \left[\left| \sigma^{-1}(p_{t,n}) - [r(\tau_{n,1}) - r(\tau_{n,0})]  \right|^2 \right]\\
    \leq & \frac{d^2}{\mu^2}\left( \frac{2 L^2 \log M}{M} + \frac{4H^2}{M^2} \right)\\
    \leq & \frac{3 d^2 L^2 \log M}{\mu^2 M},
\end{align*}
where the first inequality uses Cauchy-Schwarz inequality, and the second inequality uses Lemma.~\ref{lemma:conc-reward}. The final inequality is true by choosing $M\geq 4 (H/L)^2$. The second term comes from using the empirical average reward to estimate the value function similar to REINFORCE. So we have:
\begin{align*}
    \mathsf{Var}_{t,2} = & \E\left[\left.\left\| \frac{d}{\mu} \left(\frac{\sum_{n=1}^N \left(r(\tau_{n,1}) - r(\tau_{n,0}) \right) }{N} - \left( V(\pi_{\vtheta_t + \mu \vv_t}) - V(\pi_{\vtheta_t}) \right)\right)\vv_t \right\|_2^2 \right| \gF_t \right]\\
    = & \frac{d^2}{\mu^2 N^2} \E\left[\left.\left| \sum_{n=1}^N \left( \underbrace{\left(r(\tau_{n,1}) - r(\tau_{n,0}) \right)  - \left( V(\pi_{\vtheta_t + \mu \vv_t}) - V(\pi_{\vtheta_t}) \right)}_{\mathsf{E}_{t,n}} \right) \right|_2^2 \|\vv_t\|_2^2 \right|  \gF_t \right]\\
    = & \frac{d^2}{\mu^2 N^2} \E\left[\left.\left| \sum_{n=1}^N  \mathsf{E}_{t,n}  \right|_2^2 \right|  \gF_t \right],
\end{align*}
where the last equality is because $\vv_t$ is sampled from a unit ball. We open up the square as:
\begin{align*}
    \E\left[\left.\left| \sum_{n=1}^N  \mathsf{E}_{t,n}  \right|_2^2 \right| \gF_t \right] 
    = & \sum_{n=1}^N \E\left[\left.\left| \mathsf{E}_{t,n} \right|_2^2 \right| \gF_t \right] + \sum_{i \neq j}\E\left[ \left. \E[\mathsf{E}_{t,i} \mathsf{E}_{t,j}|\vv_t] \right| \gF_t \right]\\
    =& \sum_{n=1}^N \E\left[\left.\left| \mathsf{E}_{t,n} \right|_2^2 \right| \gF_t \right] + \sum_{i \neq j}\E\left[ \left. \E[\mathsf{E}_{t,i}|\vv_t]\right| \gF_t \right]  \E\left[ \left. \E[\mathsf{E}_{t,j}|\vv_t] \right| \gF_t \right]\\
    =& \sum_{n=1}^N \E\left[\left.\left| \mathsf{E}_{t,n} \right|_2^2 \right| \gF_t \right].
\end{align*}
where the second inequality uses the independence between trajectories generated at the same step, and the last equality is due to $\E[\left(r(\tau_{n,1}) - r(\tau_{n,0}) \right)  - \left( V(\pi_{\vtheta_t + \mu \vv_t}) - V(\pi_{\vtheta_t}) \right)] = 0$ for any fixed perturbation direction $\vv_t$ from the definition of value function. So we have:
\begin{align*}
    \mathsf{Var}_{t,2} = &  \frac{d^2}{\mu^2 N^2} \sum_{n=1}^N \E\left[\left.\left| \mathsf{E}_{t,n} \right|_2^2 \right| \gF_t \right]\\
    =& \frac{d^2}{\mu^2 N^2} \sum_{n=1}^N \E\left[\left.\left| \left(r(\tau_{n,1}) - r(\tau_{n,0}) \right)  - \left( V(\pi_{\vtheta_t + \mu \vv_t}) - V(\pi_{\vtheta_t}) \right) \right|_2^2 \right| \gF_t \right]\\
    \leq & \frac{4d^2 H^2}{\mu^2 N},
\end{align*}
where the last inequality uses the fact that both the difference of reward and the difference of value function are within $[-H, H]$. The last term to bound $\mathsf{Var}_t$ can be obtained from the property of the smoothed function in Lemma.~\ref{lemma:smoothfunction} as:
\begin{align*}
    \mathsf{Var}_{t,3} = \E\left[\left.\left\| \frac{d}{\mu}  \left( V(\pi_{\vtheta_t + \mu \vv_t}) - V(\pi_{\vtheta_t}) \right)\vv_t \right\|_2^2 \right| \gF_t \right] \leq 2d \left\| \nabla_{\vtheta} V(\pi_{\vtheta_t}) \right\|_2^2 + \frac{\mu^2 L^2 d^2}{2}.
\end{align*}
So combining three terms together, the variance drift can be bounded as:
\begin{align*}
    \mathsf{Var}_t  \leq 6d \left\| \nabla_{\vtheta} V(\pi_{\vtheta_t}) \right\|_2^2 + \frac{3\mu^2 L^2 d^2}{2} + \frac{9 d^2 L^2 \log M}{\mu^2 M} + \frac{12 d^2 H^2}{\mu^2 N}.
\end{align*}

\section{Proof of Theorem.~\ref{thm:ZBCPG}}
Due to the choice of the generation mechanism of the perturbation vector $\vv_t$, we cannot construct an explicit expression for the smoothed value function $V_\mu(\pi_{\vtheta})$ in this setting, whose gradient is somewhat "unbiased" if the correct reward function is known. In this case, we use the original value function as the Lyapunov function and perform drift analysis as follows:
\begin{align*}
    V(\pi_{\vtheta_t}) - V(\pi_{\vtheta_{t+1}}) \leq& \langle - \nabla_{\vtheta} V(\pi_{\vtheta_t}), \vtheta_{t+1} - \vtheta_t\rangle + \frac{L}{2} \|\vtheta_{t+1} - \vtheta_t\|_2^2 \\
    =& -\alpha\langle \nabla V_\mu(\pi_{\vtheta_t}), \hat{\vg}_t\rangle + \frac{L\alpha^2}{2} \|\hat{\vg}_t\|_2^2\\
    =& -\alpha\| \nabla_{\vtheta} V(\pi_{\vtheta_t}) \|_2^2 + \alpha \langle \nabla_{\vtheta} V(\pi_{\vtheta_t}), \nabla_{\vtheta} V(\pi_{\vtheta_t}) - \hat{\vg}_t\rangle + \frac{L\alpha^2}{2} \|\hat{\vg}_t\|_2^2.
\end{align*}
We also take a conditional expectation of the drift over the natural filtration $\gF_t$ of time $t$ and obtain:
\begin{equation}\label{eq:ZBCPGproof-drift}
\begin{aligned}
    &\E\left[ V(\pi_{\vtheta_t}) - V_\mu(\pi_{\vtheta_{t+1}}) | \gF_t\right]\\
    \leq & -\frac{\alpha}{2} \| \nabla_{\vtheta} V(\pi_{\vtheta_t}) \|_2^2 + \frac{L\alpha^2}{2}\underbrace{\E\left[\left.\|\hat{\vg}_t\|_2^2 \right| \gF_t \right]}_{\mathsf{Var}_t}
    + \alpha \underbrace{\langle \nabla_{\vtheta} V(\pi_{\vtheta_t}), \nabla_{\vtheta} V(\pi_{\vtheta_t}) - \E\left[\left.\hat{\vg}_t \right| \gF_t\right]\rangle}_{\mathsf{Bias}_t}.
\end{aligned}
\end{equation}
Then, we analyze the two positive drift terms $\mathsf{Bias}_t$ for the first-order gradient bias and $\mathsf{Var}_t$ for the gradient noise separately and then construct a tight upper bound of the Lyapunov drift. The results are summarized in the following two lemmas where the proofs are deferred to Sec.~\ref{sec:proofZBCPGbias} and Sec.~\ref{sec:proofZBCPGvar}:
\begin{lemma}\label{lemma:ZBCPGbias}
    For ZBCPG and let $M\geq 8 (H/L)^{\frac{2}{3}}$, conditioned on the information filtration $\gF_t$ of any time $t$, the gradient bias can be upper bounded as follows:
    \begin{align*}
        \mathsf{Bias}_t =& \langle \nabla_{\vtheta} V(\pi_{\vtheta_t}), \nabla_{\vtheta} V(\pi_{\vtheta_t}) - \E\left[\left.\hat{\vg}_t \right| \gF_t\right]\rangle
        \leq  \left(\frac{\mu Ld}{2} + \frac{2dL}{\mu}\sqrt{\frac{\log M}{M}} \right)\left\|\nabla_{\vtheta} V_\mu(\pi_{\vtheta_t}) \right\|_2.
    \end{align*}
\end{lemma}
\begin{lemma}\label{lemma:ZBCPGvar}
    For ZBCPG and let $M\geq 4 (H/L)^{2}$, conditioned on the information filtration $\gF_t$ of any time $t$, the gradient bias can be upper bounded as follows:
    \begin{align*}
        \mathsf{Var}_t =  \E\left[\left.\|\hat{\vg}_t\|_2^2 \right| \gF_t \right]  \leq 6d \left\| \nabla_{\vtheta} V(\pi_{\vtheta_t}) \right\|_2^2 + \frac{3\mu^2 L^2 d^2}{2} + \frac{9 d^2 L^2 \log M}{\mu^2 M} + \frac{12 d^2 H^2}{\mu^2 N}.
    \end{align*}
\end{lemma}
Now we use Lemma.~\ref{lemma:ZBCPGbias} and Lemma.~\ref{lemma:ZBCPGvar} and combine the drift bound in~\eqref{eq:ZBCPGproof-drift} and obtain:
\begin{align*}
    \E\left[ V(\pi_{\vtheta_t}) - V(\pi_{\vtheta_{t+1}}) | \gF_t\right]
    \leq & -\frac{\alpha}{2} \| \nabla_{\vtheta} V(\pi_{\vtheta_t}) \|_2^2  + \alpha \mathsf{Bias}_t + \frac{L\alpha^2}{2} \mathsf{Var}_t \\
    \leq & -\frac{\alpha}{2} \| \nabla_{\vtheta} V(\pi_{\vtheta_t}) \|_2^2  + \alpha \left(\frac{\mu Ld}{2} + \frac{2dL}{\mu}\sqrt{\frac{\log M}{M}} \right)\left\|\nabla_{\vtheta} V(\pi_{\vtheta_t}) \right\|_2 \\
    &+ 3\alpha^2dL \left\| \nabla_{\vtheta} V(\pi_{\vtheta_t}) \right\|_2^2 + {\alpha^2\mu^2 L^3 d^2} + \frac{5 \alpha^2 d^2 L^3 \log M}{\mu^2 M} + \frac{6 \alpha^2 d^2 H^2 L}{\mu^2 N}.
\end{align*}
Let $\alpha \leq (12dL)^{-1}$, we can further simplify the drift bound as:
\begin{align*}
    \E\left[ V(\pi_{\vtheta_t}) - V(\pi_{\vtheta_{t+1}}) | \gF_t\right] 
    \leq& -\frac{\alpha}{2} \| \nabla_{\vtheta} V(\pi_{\vtheta_t}) \|_2^2  + \alpha \left(\frac{\mu Ld}{2} + \frac{2dL}{\mu}\sqrt{\frac{\log M}{M}} \right)\left\|\nabla_{\vtheta} V(\pi_{\vtheta_t}) \right\|_2\\
    & + \frac{1}{4}\alpha \left\| \nabla_{\vtheta} V(\pi_{\vtheta_t}) \right\|_2^2 + \alpha^2\mu^2 L^3 d^2 + \frac{5 \alpha^2 d^2 L^3 \log M}{\mu^2 M} + \frac{6 \alpha^2 d^2 H^2 L}{\mu^2 N}\\
    \leq & -\frac{\alpha}{4} \| \nabla_{\vtheta} V(\pi_{\vtheta_t}) \|_2^2  + \alpha \left(\frac{\mu Ld}{2} + \frac{2dL}{\mu}\sqrt{\frac{\log M}{M}} \right)\left\|\nabla_{\vtheta} V(\pi_{\vtheta_t}) \right\|_2\\
    &+\alpha^2\mu^2 L^3 d^2 + \frac{5 \alpha^2 d^2 L^3 \log M}{\mu^2 M} + \frac{6 \alpha^2 d^2 H^2 L}{\mu^2 N},
\end{align*}
where the last inequality merges both positive and negative drifts on the square of the gradient together and assumes $d\geq 4$. When $\left\|\nabla_{\vtheta} V_\mu(\pi_{\vtheta_t}) \right\|_2 $ is large, typically, when we have:
\begin{align*}
    \left\|\nabla_{\vtheta} V_\mu(\pi_{\vtheta_t}) \right\|_2 \geq 8 \left(\frac{\mu Ld}{2} + \frac{2dL}{\mu}\sqrt{\frac{\log M}{M}} \right),
\end{align*}
we have the positive first-order drift regarding the gradient norm is bounded by:
\begin{align*}
    \left(\frac{\mu Ld}{2} + \frac{2dL}{\mu}\sqrt{\frac{\log M}{M}} \right) \left\|\nabla_{\vtheta} V_\mu(\pi_{\vtheta_t}) \right\|_2 \leq \frac{1}{8}\left\|\nabla_{\vtheta} V_\mu(\pi_{\vtheta_t}) \right\|_2^2, 
\end{align*}
and thus can be merged into the negative drift as follows:
\begin{align*}
    \E\left[ V(\pi_{\vtheta_t}) - V(\pi_{\vtheta_{t+1}}) | \gF_t\right] 
    \leq & -\frac{\alpha}{8} \| \nabla_{\vtheta} V(\pi_{\vtheta_t}) \|_2^2  +\alpha^2\mu^2 L^3 d^2 + \frac{5 \alpha^2 d^2 L^3 \log M}{\mu^2 M} + \frac{6 \alpha^2 d^2 H^2 L}{\mu^2 N}.
\end{align*}
If on the other hand, the gradient is small, i.e., when we have:
\begin{align*}
    \left\|\nabla_{\vtheta} V_\mu(\pi_{\vtheta_t}) \right\|_2 \leq 8 \left(\frac{\mu Ld}{2} + \frac{2dL}{\mu}\sqrt{\frac{\log M}{M}} \right),
\end{align*}
and then we can upper bound the first-order drift regarding the gradient norm in another way:
\begin{align*}
    \left(\frac{\mu Ld}{2} + \frac{2dL}{\mu}\sqrt{\frac{\log M}{M}} \right)\left\|\nabla_{\vtheta} V_\mu(\pi_{\vtheta_t}) \right\|_2 
    \leq& 8\left(\frac{\mu Ld}{2} + \frac{2dL}{\mu}\sqrt{\frac{\log M}{M}} \right)^2\\
    \leq & 4\mu^2 L^2d^2 + \frac{64d^2L^2 \log M}{\mu^2 M}.
\end{align*}
So in this case the total drift can be upper bounded as:
\begin{align*}
    &\E\left[ V(\pi_{\vtheta_t}) - V(\pi_{\vtheta_{t+1}}) | \gF_t\right] \\
    \leq & -\frac{\alpha}{4} \| \nabla_{\vtheta} V(\pi_{\vtheta_t}) \|_2^2  +  4\alpha\mu^2 L^2 d^2 + \alpha^2\mu^2 L^3 d^2 + \frac{64\alpha d^2L^2 \log M}{\mu^2 M}\\
    & + \frac{5 \alpha^2 d^2 L^3 \log M}{\mu^2 M} + \frac{6 \alpha^2 d^2 H^2 L}{\mu^2 N}\\
    \leq & -\frac{\alpha}{4} \| \nabla_{\vtheta} V(\pi_{\vtheta_t}) \|_2^2   + 5\alpha\mu^2 L^2 d^2 + \frac{64\alpha d^2L^2 \log M}{\mu^2 M} + \frac{5 \alpha^2 d^2 L^3 \log M}{\mu^2 M} + \frac{6 \alpha^2 d^2 H^2 L}{\mu^2 N}.
\end{align*}
The last inequality uses $\alpha \leq (12dL)^{-1}$ and $d\geq4$. So combining both bounds and taking an expectation, we have:
\begin{align*}
    &\E\left[ V(\pi_{\vtheta_t}) - V(\pi_{\vtheta_{t+1}}) | \gF_t\right] \\
    \leq & -\frac{\alpha}{8} \E\left[ \| \nabla_{\vtheta} V(\pi_{\vtheta_t}) \|_2^2\right]  + 5\alpha\mu^2 L^2 d^2 + \frac{5 \alpha^2 d^2 L^3 \log M}{\mu^2 M} + \frac{64\alpha d^2L^2 \log M}{\mu^2 M} + \frac{6 \alpha^2 d^2 H^2 L}{\mu^2 N}\\
    \leq & -\frac{\alpha}{8} \E\left[ \| \nabla_{\vtheta} V(\pi_{\vtheta_t}) \|_2^2\right]  + 5\alpha\mu^2 L^2 d^2 + \frac{65\alpha d^2L^2 \log M}{\mu^2 M} + \frac{6 \alpha^2 d^2 H^2 L}{\mu^2 N},
\end{align*}
where we use $\alpha \leq (12dL)^{-1}$ again the in the last inequality. We then take a telescoping sum which results in the following:
\begin{align*}
    &\E\left[ V_\mu(\pi_{\vtheta_0})\right] - \E\left[ V_\mu(\pi_{\vtheta_T})\right]\\
    \leq& -\frac{\alpha}{8} \sum_{t=0}^{T-1}\E\left[ \| \nabla_{\vtheta} V(\pi_{\vtheta_t}) \|_2^2\right]  + \left(5\alpha\mu^2 L^2 d^2 + \frac{65\alpha d^2L^2 \log M}{\mu^2 M} + \frac{6 \alpha^2 d^2 H^2 L}{\mu^2 N} \right)T.
\end{align*}
We choose $\mu$ to balance the terms inside the parenthesis. Specifically, let
\begin{align*}
    \mu^2 = \max\left\{ 4\sqrt{\frac{\log M}{M}}, \frac{H}{3L\sqrt{dN}} \right\}, \quad \alpha = \frac{1}{12 d L}.
\end{align*}
Then, we have the following inequality on the positive drift:
\begin{align*}
    5\alpha\mu^2 L^2 d^2 \geq \frac{65\alpha d^2L^2 \log M}{\mu^2 M}, \quad 5\alpha\mu^2 L^2 d^2 \geq \frac{6 \alpha^2 d^2 H^2 L}{\mu^2 N}.
\end{align*}
So we have the final results in Theorem.~\ref{thm:ZBCPG} as:
\begin{align*}
    \frac{1}{T}\sum_{t=0}^{T-1}\E\left[ \| \nabla_{\vtheta} V(\pi_{\vtheta_t}) \|_2^2\right] 
    \leq & \frac{8\left( \E\left[ V(\pi_{\vtheta_0})\right] - \E\left[ V(\pi_{\vtheta_T})\right] \right)}{T\alpha} + 120\mu^2 L^2 d^2\\
    =& \gO\left( \frac{H Ld}{T} + \frac{d^2 L^2 \sqrt{\log M}}{\sqrt{M}} + \frac{HL d\sqrt{d}}{\sqrt{N}} \right).
\end{align*}

\subsection{Proof of Lemma.~\ref{lemma:ZBCPGbias}}\label{sec:proofZBCPGbias}

We analyze the first-order gradient bias term $\mathsf{Bias}_t$ in the drift of~\eqref{eq:ZBCPGproof-drift}. Notice that conditioned over $\gF_t$, the randomness only comes from sampling the perturbation coordinate set $\vi_t$, the perturbation direction $\boldsymbol{\lambda}_t$, sampling trajectories, and obtaining human feedback. By~\eqref{eq:smooth-grad-def} and the definition of $\hat{\vg}_t$, we have:
\begin{align*}
    \nabla_{\vtheta} V(\pi_{\vtheta_t}) - \E\left[ \left.\hat{\vg}_t \right| \gF_t\right]  
    =& \nabla_{\vtheta} V(\pi_{\vtheta_t}) - \E \left[\left. \E\left[ \left.\frac{d}{ \mu }\frac{\sum_{n=1}^N \sigma^{-1}(p_{t,n})}{N} \right| \vv_t\right] \vv_t \right| \gF_t\right]\\
    =& \nabla_{\vtheta} V(\pi_{\vtheta_t}) - \frac{d}{ \mu } \E\left[ \left. \E\left[\left. \sigma^{-1}(p_{t,n}) \right| \vv_t\right]\vv_t \right| \gF_t\right]\\
    =& \underbrace{\nabla_{\vtheta} V(\pi_{\vtheta_t}) - \E\left[\left. \frac{d}{\mu }\left( V(\pi_{\vtheta_t + \mu \vv_t}) - V(\pi_{\vtheta_t}) \right) \vv_t  \right| \gF_t\right]}_{\mathsf{Bias_{t,1}}}\\
    & + \frac{d}{ \mu } \E\left[\left. \underbrace{\left[ \left(V(\pi_{\vtheta_t + \mu\vv_t}) - V(\pi_{\vtheta_t})\right) - \E\left[\left. r(\tau_{n,1}) - r(\tau_{n,0}) \right| \vv_t \right] \right]}_{=0}  \vv_t \right| \gF_t \right] \\
    &+ \underbrace{\frac{d}{ \mu }\E\left[\left. \E\left[\left. \left( \sigma^{-1}(p_{t,n}) - [r(\tau_{n,1}) - r(\tau_{n,0})] \right) \right| \tau_{n,1}, \tau_{n,0} \right] \vv_t \right| \gF_t \right]}_{\mathsf{Bias}_{t,2}},
\end{align*}
where the second inequality uses the fact that conditioned on $\vv_t$, each trajectory pair $(\tau_{n,1}, \tau_{n,0})$ is independently generated from policies $(\pi_{\vtheta_t + \mu\vv_t}, \pi_{\vtheta_t})$. The last inequality is since conditioned on the policy $(\pi_{\vtheta_t + \mu\vv_t}, \pi_{\vtheta_t})$, the expectation of each reward trajectory is the value function of the policy since $(\tau_{n,1}, \tau_{n,0})$ is independently generated from policies $(\pi_{\vtheta_t + \mu\vv_t}, \pi_{\vtheta_t})$. Therefore, the $\mathsf{Bias}_t$ term is written as follows:
\begin{align*}
    \mathsf{Bias_t} = \langle \nabla_{\vtheta} V(\pi_{\vtheta_t}), \mathsf{Bias}_{t,1}\rangle + \langle \nabla_{\vtheta} V(\pi_{\vtheta_t}), \mathsf{Bias}_{t,2}\rangle.
\end{align*}
We first analyze the second term which comes from preference estimation. Similar to the proof of ZPG, we have:
\begin{align*}
    \langle \nabla_{\vtheta} V(\pi_{\vtheta_t}), \mathsf{Bias}_{t,2}\rangle 
    = & \frac{d}{ \mu }\E\left[\left. \E\left[\left. \left( \sigma^{-1}(p_{t,n}) - [r(\tau_{n,1}) - r(\tau_{n,0})] \right) \right| \tau_{n,1}, \tau_{n,0} \right] \langle \nabla_{\vtheta} V(\pi_{\vtheta_t}) , \vv_t \rangle \right| \gF_t \right]\\
    \leq & \frac{d}{\mu }\|\nabla_{\vtheta} V(\pi_{\vtheta_t})\|_2  \E\left[\left. \E\left[\left. \left|\sigma^{-1}(p_{t,n}) - \left[r(\tau_{n,1}) - r(\tau_{n,0})\right] \right| \right| \tau_{n,1}, \tau_{n,0} \right]  \right| \gF_t \right]\\
    \leq & \frac{d}{\mu }\|\nabla_{\vtheta} V(\pi_{\vtheta_t})\|_2 \left(L\sqrt{\frac{2\log M}{M}} + \frac{2H}{M^2}\right)\\
    \leq& \frac{2dL}{\mu}\sqrt{\frac{\log M}{M}} \left\|\nabla_{\vtheta} V_\mu(\pi_{\vtheta_t}) \right\|_2 ,
\end{align*}
where the first inequality uses Cauchy-Schwarz inequality and $\|\vv_t\|_2 = 1$ and the second inequality uses Lemma.~\ref{lemma:conc-reward}. The last inequality is obtained by choosing $M \geq 8 (H/L)^{\frac{2}{3}}$. Then, we need to deal with the first-term inner product. We first make the following claim:
\begin{lemma}
    For any iteration $t$, the following is true:
    \begin{align*}
    \frac{1}{d} \nabla_{\vtheta} V(\pi_{\vtheta_t}) = \E_{\vi_t ,\boldsymbol{\lambda}_t}\left[ \langle \nabla_{\vtheta} V(\pi_{\vtheta_t}), \vv_t\rangle \vv_t\right],
\end{align*}
where the subscript indicates the randomness only comes from $\vi_t $ and $\boldsymbol{\lambda}_t$.
\end{lemma}
\textbf{Proof.} To prove this equality, we start from the right-hand side:
\begin{align*}
    \E_{\vi_t ,\boldsymbol{\lambda}_t}\left[ \langle \nabla_{\vtheta} V(\pi_{\vtheta_t}), \vv_t\rangle \vv_t\right] 
    = & \frac{1}{K}\E_{\vi_t ,\boldsymbol{\lambda}_t}\left[ \left\langle \nabla_{\vtheta} V(\pi_{\vtheta_t}), \sum_{j=1}^K \lambda_{i_{t,j}} \ve_{i_{t,j}} \right\rangle \sum_{l=1}^K \lambda_{i_{t,l}} \ve_{i_{t,l}}\right]\\
    =& \frac{1}{K}\E_{\vi_t ,\boldsymbol{\lambda}_t}\left[ \sum_{j=1}^K \sum_{l=1}^K \left\langle \nabla_{\vtheta} V(\pi_{\vtheta_t}),  \lambda_{i_{t,j}} \ve_{i_{t,j}} \right\rangle \lambda_{i_{t,l}} \ve_{i_{t,l}}\right]\\
    =& \frac{1}{K}\E_{\vi_t}\left[ \sum_{j=1}^K \sum_{l=1}^K \left\langle \nabla_{\vtheta} V(\pi_{\vtheta_t}),   \ve_{i_{t,j}} \right\rangle  \ve_{i_{t,l}} \E_{\boldsymbol{\lambda}_t}[\lambda_{i_{t,j}}\lambda_{i_{t,l}}] \right],
\end{align*}
where the first equality is by the definition of $\vv_t$, and the third equality is by the law of total expectation. For $j\neq l $, we can use the independence between different coordinates of $\boldsymbol{\lambda}_t$. Since the expectation of $\boldsymbol{\lambda}_t$ is zero, we have:
\begin{align*}
    \E_{\boldsymbol{\lambda}_t}[\lambda_{i_{t,j}}\lambda_{i_{t,l}}] = \E_{\boldsymbol{\lambda}_t}[\lambda_{i_{t,j}}]\E_{\boldsymbol{\lambda}_t}[\lambda_{i_{t,l}}] = 0.
\end{align*}
On the other hand, if $j = l$, since $\lambda_{i_t,j}$ is sampled from $\{-1, 1\}$, we have $\lambda_{i_t,j}^2 = 1$. Therefore, the right-hand side can be expressed as:
\begin{align*}
    \E_{\vi_t ,\boldsymbol{\lambda}_t}\left[ \langle \nabla_{\vtheta} V(\pi_{\vtheta_t}), \vv_t\rangle \vv_t\right] =& \frac{1}{K}\E_{\vi_t}\left[ \sum_{j=1}^K \left\langle \nabla_{\vtheta} V(\pi_{\vtheta_t}),   \ve_{i_{t,j}} \right\rangle  \ve_{i_{t,j}}\right]\\
    =& \frac{1}{K}\E_{\vi_t}\left[ \sum_{i=1}^d \mathbbm{1}_{i \in \vi_t}\left\langle \nabla_{\vtheta} V(\pi_{\vtheta_t}),   \ve_{i} \right\rangle  \ve_{i}\right]\\
    =&  \frac{1}{K}\sum_{i=1}^d \sP(i \in \vi_t)\left\langle \nabla_{\vtheta} V(\pi_{\vtheta_t}),   \ve_{i} \right\rangle  \ve_{i}\\
    =& \frac{1}{d}\sum_{i=1}^d \left\langle \nabla_{\vtheta} V(\pi_{\vtheta_t}),   \ve_{i} \right\rangle  \ve_{i}\\
    =& \frac{1}{d} \nabla_{\vtheta} V(\pi_{\vtheta_t}),
\end{align*}
where the second last equality is because the probability of each coordinate is chosen by $\vi_t$ is exactly $K/d$. The proof is hence concluded. $\hfill\blacksquare$

With the help of the Lemma above, we can express the $\mathsf{Bias}_{t,1}$ term as follows:
\begin{align*}
    \langle \nabla_{\vtheta} V(\pi_{\vtheta_t}), \mathsf{Bias}_{t,1}\rangle
    =& \left\langle \nabla_{\vtheta} V(\pi_{\vtheta_t}), \nabla_{\vtheta} V(\pi_{\vtheta_t}) - \frac{d}{\mu}\E\left[\left. \left( V(\pi_{\vtheta_t + \mu \vv_t}) - V(\pi_{\vtheta_t}) \right) \vv_t  \right| \gF_t\right] \right\rangle\\
    =& \frac{d}{\mu}\left\langle \nabla_{\vtheta} V(\pi_{\vtheta_t}), \E\left[ \frac{\mu}{d}\nabla_{\vtheta} V(\pi_{\vtheta_t}) - \left. \left( V(\pi_{\vtheta_t + \mu \vv_t}) - V(\pi_{\vtheta_t}) \right) \vv_t  \right| \gF_t\right] \right\rangle \\
    =& - \frac{d}{\mu}\E\left[\left. \left( V(\pi_{\vtheta_t + \mu \vv_t}) - V(\pi_{\vtheta_t}) -\langle \nabla_{\vtheta} V(\pi_{\vtheta_t}), \mu\vv_t\rangle \right) \left\langle\nabla_{\vtheta} V(\pi_{\vtheta_t}), \vv_t \right\rangle \right| \gF_t\right]\\
    \leq & \frac{d}{\mu}\E\left[\left. \left| V(\pi_{\vtheta_t + \mu \vv_t}) - V(\pi_{\vtheta_t}) -\langle \nabla_{\vtheta} V(\pi_{\vtheta_t}), \mu\vv_t\rangle \right| \left\|\nabla_{\vtheta} V(\pi_{\vtheta_t}) \right\|_2 \left\|\vv_t \right\|_2 \right| \gF_t\right].
\end{align*}
By the smoothness of the value function $V$, we have:
\begin{align*}
    \left| V(\pi_{\vtheta_t + \mu \vv_t}) - V(\pi_{\vtheta_t}) -\langle \nabla_{\vtheta} V(\pi_{\vtheta_t}), \mu\vv_t\rangle \right| \leq \frac{\mu^2 L}{2}\|\vv_t\|_2 = \frac{\mu^2 L}{2}.
\end{align*}
So we can substitute into the bias term and obtain:
\begin{align*}
    \langle \nabla_{\vtheta} V(\pi_{\vtheta_t}), \mathsf{Bias}_{t,1}\rangle
    \leq & \frac{d}{\mu}\E\left[\left. \frac{\mu^2 L}{2} \left\|\nabla_{\vtheta} V(\pi_{\vtheta_t}) \right\|_2 \right| \gF_t\right]
    = \frac{\mu Ld}{2} \left\|\nabla_{\vtheta} V(\pi_{\vtheta_t}) \right\|_2.
\end{align*}
Then, we obtain the upper bound of the gradient bias as follows:
\begin{align*}
    \mathsf{Bias_t} = \langle \nabla_{\vtheta} V(\pi_{\vtheta_t}), \mathsf{Bias}_{t,1} + \mathsf{Bias}_{t,2}\rangle
    \leq \left(\frac{\mu Ld}{2} + \frac{2dL}{\mu}\sqrt{\frac{\log M}{M}} \right)\left\|\nabla_{\vtheta} V_\mu(\pi_{\vtheta_t}) \right\|_2.
\end{align*}

\subsection{Proof of Lemma.~\ref{lemma:ZBCPGvar}}\label{sec:proofZBCPGvar}
We aim for the variance term $\mathsf{Var}_t$ in the Lyapunov drift in~\eqref{eq:ZBCPGproof-drift}. We first have upper bound variance as follows:
\begin{align*}
    \mathsf{Var}_t 
    = & \E\left[\left.\left\|\frac{d}{ \mu }\frac{\sum_{n=1}^N \sigma^{-1}(p_{t,n})}{N} \vv_t \right\|_2^2 \right| \gF_t \right]\\
    \leq & 3 \underbrace{\E\left[\left.\left\|\frac{d \sum_{n=1}^N \left(\sigma^{-1}(p_{t,n}) - [r(\tau_{n,1}) - r(\tau_{n,0})] \right) }{ \mu N}\vv_t \right\|_2^2 \right| \gF_t \right]}_{\mathsf{Var}_{t,1}}\\
    &+ 3 \underbrace{\E\left[\left.\left\| \frac{d}{ \mu} \left(\frac{\sum_{n=1}^N \left(r(\tau_{n,1}) - r(\tau_{n,0}) \right) }{N} - \left( V(\pi_{\vtheta_t + \mu \vv_t}) - V(\pi_{\vtheta_t}) \right)\right)\vv_t \right\|_2^2 \right| \gF_t \right]}_{\mathsf{Var}_{t,2}}\\
    & + 3\underbrace{\E\left[\left.\left\| \frac{d}{\mu}  \left( V(\pi_{\vtheta_t + \mu \vv_t}) - V(\pi_{\vtheta_t}) \right)\vv_t \right\|_2^2 \right| \gF_t \right]}_{\mathsf{Var}_{t,3}}.
\end{align*}
The first term comes from the human preference estimation error which can be bounded using \eqref{eq:conc-reward2} from Lemma.~\ref{lemma:conc-reward} as follows:
\begin{align*}
    \mathsf{Var}_{t,1} 
    = &\E\left[\left.\left\|\frac{d \sum_{n=1}^N \left(\sigma^{-1}(p_{t,n}) - [r(\tau_{n,1}) - r(\tau_{n,0})] \right) }{ \mu N}\vv_t \right\|_2^2 \right| \gF_t \right]\\
    = & \frac{d^2}{\mu^2 N^2} \E \left[\left. \left|\sum_{n=1}^N \left(\sigma^{-1}(p_{t,n}) - [r(\tau_{n,1}) - r(\tau_{n,0})] \right) \right|^2 \|\vv_t\|_2^2 \right|\gF_t\right]\\
    \leq & \frac{d^2}{\mu^2 N} \sum_{n=1}^N \E \left[\left| \sigma^{-1}(p_{t,n}) - [r(\tau_{n,1}) - r(\tau_{n,0})]  \right|^2 \right]\\
    \leq & \frac{d^2}{\mu^2}\left( \frac{2 L^2 \log M}{M} + \frac{4H^2}{M^2} \right)\\
    \leq & \frac{3 d^2 L^2 \log M}{\mu^2 M},
\end{align*}
where the first inequality uses Cauchy-Schwarz inequality and the fact that $\|\vv_t\|_2^2 = 1$, and the second inequality uses Lemma.~\ref{lemma:conc-reward}. The final inequality is true by choosing $M\geq 4 (H/L)^2$. The second term comes from using the empirical average reward to estimate the value function similar to REINFORCE. So we have:
\begin{align*}
    \mathsf{Var}_{t,2} = & \E\left[\left.\left\| \frac{d}{\mu} \left(\frac{\sum_{n=1}^N \left(r(\tau_{n,1}) - r(\tau_{n,0}) \right) }{N} - \left( V(\pi_{\vtheta_t + \mu \vv_t}) - V(\pi_{\vtheta_t}) \right)\right)\vv_t \right\|_2^2 \right| \gF_t \right]\\
    = & \frac{d^2}{\mu^2 N^2} \E\left[\left.\left| \sum_{n=1}^N \left( \underbrace{\left(r(\tau_{n,1}) - r(\tau_{n,0}) \right)  - \left( V(\pi_{\vtheta_t + \mu \vv_t}) - V(\pi_{\vtheta_t}) \right)}_{\mathsf{E}_{t,n}} \right) \right|_2^2 \|\vv_t\|_2^2 \right|  \gF_t \right]\\
    = & \frac{d^2}{\mu^2 N^2} \E\left[\left.\left| \sum_{n=1}^N  \mathsf{E}_{t,n}  \right|_2^2 \right|  \gF_t \right],
\end{align*}
where the last equality is because $\|\vv_t\|_2^2 = 1$. We open up the square as:
\begin{align*}
    \E\left[\left.\left| \sum_{n=1}^N  \mathsf{E}_{t,n}  \right|_2^2 \right| \gF_t \right] 
    = & \sum_{n=1}^N \E\left[\left.\left| \mathsf{E}_{t,n} \right|_2^2 \right| \gF_t \right] + \sum_{i \neq j}\E\left[ \left. \E[\mathsf{E}_{t,i} \mathsf{E}_{t,j}|\vv_t] \right| \gF_t \right]\\
    =& \sum_{n=1}^N \E\left[\left.\left| \mathsf{E}_{t,n} \right|_2^2 \right| \gF_t \right] + \sum_{i \neq j}\E\left[ \left. \E[\mathsf{E}_{t,i}|\vv_t]\right| \gF_t \right]  \E\left[ \left. \E[\mathsf{E}_{t,j}|\vv_t] \right| \gF_t \right]\\
    =& \sum_{n=1}^N \E\left[\left.\left| \mathsf{E}_{t,n} \right|_2^2 \right| \gF_t \right].
\end{align*}
where the second inequality uses the independence between trajectories generated at the same step, and the last equality is due to $\E[\left(r(\tau_{n,1}) - r(\tau_{n,0}) \right)  - \left( V(\pi_{\vtheta_t + \mu \vv_t}) - V(\pi_{\vtheta_t}) \right)] = 0$ for any fixed perturbation direction $\vv_t$ from the definition of value function. So we have:
\begin{align*}
    \mathsf{Var}_{t,2} = &  \frac{d^2}{\mu^2 N^2} \sum_{n=1}^N \E\left[\left.\left| \mathsf{E}_{t,n} \right|_2^2 \right| \gF_t \right]\\
    =& \frac{d^2}{\mu^2 N^2} \sum_{n=1}^N \E\left[\left.\left| \left(r(\tau_{n,1}) - r(\tau_{n,0}) \right)  - \left( V(\pi_{\vtheta_t + \mu \vv_t}) - V(\pi_{\vtheta_t}) \right) \right|_2^2 \right| \gF_t \right]\\
    \leq & \frac{4d^2 H^2}{\mu^2 N},
\end{align*}
where the last inequality uses the fact that both the difference of reward and the difference of value function are within $[-H, H]$. Now we analyze the last term:
\begin{align*}
    \mathsf{Var}_{t,3} 
    = & \E\left[\left.\left\| \frac{d}{\mu}  \left( V(\pi_{\vtheta_t + \mu \vv_t}) - V(\pi_{\vtheta_t}) \right)\vv_t \right\|_2^2 \right| \gF_t \right] \\
    = & \frac{d^2}{\mu^2} \E\left[\left.\left\| \left( V(\pi_{\vtheta_t + \mu \vv_t}) - V(\pi_{\vtheta_t}) - \langle \nabla_{\vtheta} V(\pi_{\vtheta_t}), \mu \vv_t \rangle \right)\vv_t + \langle \nabla_{\vtheta} V(\pi_{\vtheta_t}), \mu \vv_t \rangle\vv_t \right\|_2^2 \right| \gF_t \right]\\
    \leq & \frac{2d^2}{\mu^2} \E\left[\left.\left\| \left( V(\pi_{\vtheta_t + \mu \vv_t}) - V(\pi_{\vtheta_t}) - \langle \nabla_{\vtheta} V(\pi_{\vtheta_t}), \mu \vv_t \rangle \right)\vv_t\right\|_2^2 \right| \gF_t \right]
    + \frac{2d^2}{\mu^2} \E\left[\left.\left\| \langle \nabla_{\vtheta} V(\pi_{\vtheta_t}), \mu \vv_t \rangle\vv_t \right\|_2^2 \right| \gF_t \right].
\end{align*}
The first term can be bounded using the smoothness of the value function as follows:
\begin{align*}
    &\E\left[\left.\left\| \left( V(\pi_{\vtheta_t + \mu \vv_t}) - V(\pi_{\vtheta_t}) - \langle \nabla_{\vtheta} V(\pi_{\vtheta_t}), \mu \vv_t \rangle \right)\vv_t\right\|_2^2 \right| \gF_t \right]\\
    = &\E\left[\left.\left| \left( V(\pi_{\vtheta_t + \mu \vv_t}) - V(\pi_{\vtheta_t}) - \langle \nabla_{\vtheta} V(\pi_{\vtheta_t}), \mu \vv_t \rangle \right)\right|_2^2 \|\vv_t\|_2^2 \right| \gF_t \right]\\
    \leq & \E\left[\left.\frac{L^2\mu^4}{4} \|\vv_t\|_2^4 \right| \gF_t \right]\\
    =& \frac{L^2\mu^4}{4},
\end{align*}
where the last equality uses $\|\vv_t\|_2^2 = K$. The second term can be simplified as follows:
\begin{align*}
    \E\left[\left.\left\| \langle \nabla_{\vtheta} V(\pi_{\vtheta_t}), \mu \vv_t \rangle\vv_t \right\|_2^2 \right| \gF_t \right] 
    = & \mu^2 \E\left[\left. \nabla_{\vtheta} V(\pi_{\vtheta_t})^{\top} \vv_t  \vv_t^{\top} \nabla_{\vtheta} V(\pi_{\vtheta_t})\left\| \vv_t \right\|_2^2 \right| \gF_t \right]\\
    = & \mu^2 \nabla_{\vtheta} V(\pi_{\vtheta_t})^{\top} \E\left[\left. \vv_t  \vv_t^{\top} \right| \gF_t \right]\nabla_{\vtheta} V(\pi_{\vtheta_t}).
\end{align*}
Notice that $\vv_t$ is a vector such that on the coordinates chosen by $\vi_t$, the element is either $-1/\sqrt{K}$ or $1/\sqrt{K}$, and on other coordinates, it is zero. So $\vv_t \vv_t^{\top}$ will result in a diagonal matrix where on position $(i_{t,1}, i_{t,1}), \cdots, (i_{t,K},i_{t,K})$ the element is $1/K$ and $0$ otherwise. Since each coordinate $i$ will be chosen in each iteration in $\vi_t$ with probability $K/d$, taking an expectation over $\vv_t \vv_t^{\top}$ will result in a diagonal matrix where the diagonal elements are $1/d$. Therefore, we have:
\begin{align*}
    \E\left[\left.\left\| \langle \nabla_{\vtheta} V(\pi_{\vtheta_t}), \mu \vv_t \rangle\vv_t \right\|_2^2 \right| \gF_t \right] 
    = & \frac{\mu^2}{d} \nabla_{\vtheta} V(\pi_{\vtheta_t})^{\top} \nabla_{\vtheta} V(\pi_{\vtheta_t}) = \frac{\mu^2}{d} \|\nabla_{\vtheta} V(\pi_{\vtheta_t})\|_2^2.
\end{align*}
So Let us combine both terms and we can bound $\mathsf{Var}_{t,3}$ as follows:
\begin{align*}
    \mathsf{Var}_{t,3}  \leq \frac{\mu^2L^2d^2}{2} + 2d\|\nabla_{\vtheta} V(\pi_{\vtheta_t})\|_2^2.
\end{align*}
So combining three terms, the variance drift can be bounded as:
\begin{align*}
    \mathsf{Var}_t  \leq 6d \left\| \nabla_{\vtheta} V(\pi_{\vtheta_t}) \right\|_2^2 + \frac{3\mu^2 L^2 d^2}{2} + \frac{9 d^2 L^2 \log M}{\mu^2 M} + \frac{12 d^2 H^2}{\mu^2 N}.
\end{align*}

\end{document}